\documentclass[sigconf, nonacm]{acmart}
\AtBeginDocument{%
  }

\setcopyright{acmlicensed}
\copyrightyear{2018}
\acmYear{2018}
\acmDOI{XXXXXXX.XXXXXXX}
\acmConference[KDD '26]{Make sure to enter the correct
  conference title from your rights confirmation email}{August 09--13,
  2026}{Jeju, Korea}
\acmISBN{978-1-4503-XXXX-X/2018/06}

\DeclareMathOperator*{\argmax}{arg\,max}

\usepackage{multirow}
\usepackage{subcaption}
\usepackage{listings}
\begin{document}

\title{Managing Map Cardinality in Automatic Disease Classification Mapping: Balancing Precision, Recall and Coverage}

\author{Santosh Purja Pun}
\authornote{Corresponding Author}
\email{18870679@student.westernsydney.edu.au}
\orcid{0000-0002-9820-2553}
\affiliation{%
  \institution{Western Sydney University}
  \city{Parramatta}
  \state{NSW}
  \country{Australia}
}

\author{Oliver Obst}
\email{o.obst@unsw.edu.au}
\orcid{0000-0002-8284-2062}
\affiliation{%
  \institution{UNSW}
  \city{Sydney}
  \state{NSW}
  \country{Australia}}

\author {Jim Basilakis}
\email{j.basilakis@westernsydney.edu.au}
\orcid{0000-0002-7440-1320}
\affiliation{%
  \institution{Western Sydney University}
  \city{Parramatta}
  \state{NSW}
  \country{Australia}
}

\author{Jeewani Anupama Ginige}
\email{j.ginige@westernsydney.edu.au}
\orcid{0000-0002-6695-6983}
\affiliation{%
 \institution{Western Sydney University}
  \city{Parramatta}
  \state{NSW}
  \country{Australia}}

\renewcommand{\shortauthors}{S Purja Pun et al.}

\begin{abstract}
  Automatic mapping between disease classification systems, such as the International Classification of Diseases (ICD), is a challenging yet essential task for integrating health data and conducting longitudinal data analysis. Existing embedding-based methods primarily focus on \emph{one-to-one} mappings, overlooking more complex \emph{one-to-many} scenarios. The threshold-based and top-K methods offer natural extensions; however, they involve inherent trade-offs between \emph{precision}, \emph{recall} and \emph{mapping coverage}---the proportion of source codes with at least one mapping to a target code. To address this challenge, we introduce a novel method, which is inspired by the \emph{blocking-and-matching} pipeline commonly used in \emph{entity resolution}. In particular, we first generate a block of candidate matches (\emph{blocking}) and then employ a large language model (LLM) to identify all valid mappings within each block (\emph{matching}). Empirically, we show that the proposed method achieves higher precision with comparable recall and broader coverage across multiple ICD version pairs (ICD-9-CM$\leftrightarrow$ICD-10-CM and ICD-10-AM$\leftrightarrow$ICD-11). Our source code and dataset is available at: \url{https://tinyurl.com/46kyn7wp}.
\end{abstract}

\ccsdesc[500]{Applied computing~Health informatics}

\keywords{Mapping Clinical Systems, Clinical Terminology Alignment, Automatic Mapping of Disease Classification System}
\begin{teaserfigure}
  \includegraphics[width=\linewidth]{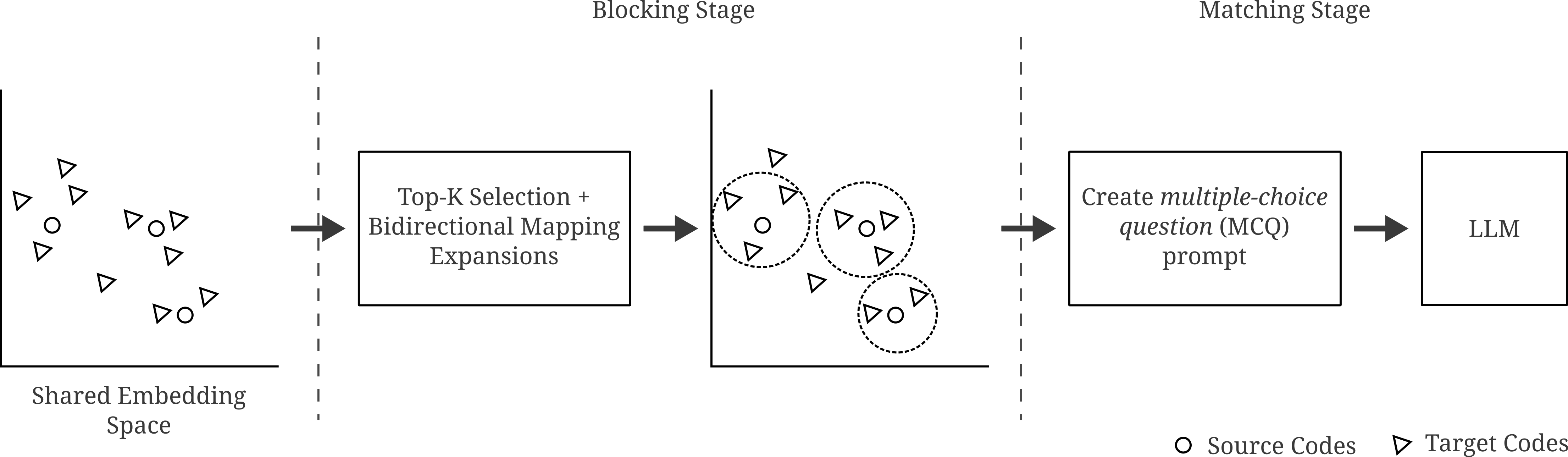}
    \caption{Illustration of our automatic mapping method, inspired by the \emph{blocking-and-matching} pipeline commonly used in \emph{entity resolution}. It consists of two stages: (1) a blocking stage, which employs an embedding-based selection method to create a set of potential matches, and (2) a matching stage, which formulates the matching task as a \emph{multiple-choice question} (MCQ) and uses LLM as a selector. }
    \label{fig:mapping-pipeline}
    \Description{Illustration of the proposed blocking-and-matching framework for automatic mapping of disease classification systems.}
\end{teaserfigure}


\maketitle

\section{Introduction}
Disease classification systems, such as the International Classification of Diseases (ICD), provide standardized coding schemas to capture, analyze, and report data from clinical documents, including discharge summaries and clinical notes. These systems evolve to reflect recent advances in medical science. For instance, over time, ICD has evolved through revisions such as ICD-9 and ICD-10 to ICD-11, the current version. Such evolution creates a fundamental problem in health data integration and interoperability: different ICD versions use different coding standards, often with conditions defined at varying levels of granularity. Additionally, different countries may extend the base version into country-specific adaptations, such as CM (Clinical Modification) in the US and AM (Australian Modification) in Australia, adding another layer of complexity when comparing health data across countries.

Mapping tables provide a crosswalk between different versions; however, they are created manually by terminology experts, which is both time-consuming and labour-intensive, and offer little to no scalability. Automatic mapping approaches address some of these limitations of the manual process, but progress has been minimal. While earlier methods rely on simple string comparison or query extension techniques~\citep{allones2014automated, huang2009using, wang2008computational}, recent embedding-based methods primarily focus on \emph{one-to-one} mappings (for examples see our previous works~\citep{10.1093/jamia/ocag004, purja2025managing}), excluding more challenging and clinically significant \emph{one-to-many} mappings.

\emph{One-to-many} mappings in ICD arise when a source concept is semantically broader than any individual target concept, and a combination of multiple target codes more accurately reflects its meaning. For example, the ICD-9-CM code \textbf{065.8} (\emph{Mosquito-borne hemorrhagic fever}) maps to both ICD-10-CM code \textbf{A91} (\emph{Dengue hemorrhagic fever}) and \textbf{A92.0} (\emph{Chikungunya virus disease}), as these together represent the range of conditions described by the broader source code. These cases are inherently more challenging than \emph{one-to-one} mappings, as they require a more nuanced semantic understanding beyond simple pairwise similarity.

One straightforward approach to handling the \emph{cardinality of the maps} is to extend embedding-based methods using selection methods such as a \emph{threshold-based} and \emph{top-K}. While these methods are simple and provide a natural extension to embedding-based methods, they suffer from inherent trade-offs between precision and recall. For instance, with higher threshold values, the threshold-based method prioritizes precision at the expense of recall, whereas lower values prioritize recall over precision. The top-$K$ selection method, on the other hand, generally optimizes recall over precision, resulting in a high number of false positives.

Additionally, the mapping coverage, which is the proportion of source code with at least one mapping to a target code, adds another complexity to the precision-recall trade-off. For instance, in threshold-based selection, higher thresholds yield lower mapping coverage, whereas lower values yield higher coverage. While selecting $K$ target codes with the highest similarity scores (top-$K$) would ideally solve the mapping coverage problem, it would sacrifice precision by including many false positives. For instance, for valid $1:1$ mappings (which constitute the majority of ICD mappings),  the top-$K$ generates $K-1$ false positives per source code. 

These trade-offs have critical practical implications. As outlined in the WHO-FIC Classification and Terminology Mapping~\citep{world2021fic}, the generated maps must undergo an independent manual quality assurance and usage validation process. As such, these approaches present terminology experts with one of two unfavourable scenarios: 

\begin{enumerate}
	\item{High-recall, low-precision methods (e.g., Top-K) generate excessive  false positives, overwhelming validators with irrelevant candidates}
	\item{High-precision, low-coverage methods (e.g., threshold-based) leave  large gaps requiring manual mapping from scratch }
\end{enumerate}

Thus, an ideal mapping system must generate high-precision maps with comparable recall and broader coverage to assist the manual validation process by reducing manual workloads.

To this end, we propose a novel method, inspired by the \emph{blocking-and-matching} pipeline commonly used in \emph{entity resolution} (ER), for automatically generating mappings between different ICD versions (see Figure~\ref{fig:mapping-pipeline}). Our approach consists of two main stages: (1) a \emph{blocking} stage, in which we generate a block (i.e., a set) of candidate target codes for each source code; and (2) a \emph{matching} stage, in which we use a large language model (LLM) to identify all matching target codes. For the blocking stage, we introduce a hybrid embedding-based selection strategy that combines the top-K selection and \emph{bidirectional mapping expansion} (BiMaps). Likewise, in the matching phase, we reformulate the matching problem as a multiple-choice question (MCQ) and prompt the LLM to select all matching options (if any).

Empirically, we demonstrate the effectiveness of our method by mapping pairs of ICD versions, namely ICD-9-CM$\leftrightarrow$ICD-10-CM and ICD-10-AM$\leftrightarrow$ICD-11. Given that the ICD versions organize diseases by anatomical structures into different chapters, we adopt a \emph{chapter-wise} mapping approach that maps codes between equivalent chapters of the source and target ICD versions. While we acknowledge that some source codes may map to target codes in different chapters, such cases are typically few. To account for this, we adjust the ground truth: we exclude those source codes that map completely to codes in different chapters in the target system. For partially mapped codes, we take the in-chapter target codes as the complete mappings.

Our main contributions are as follows:
\begin{enumerate}
    \item We introduce a \emph{blocking‑and‑matching} framework that combines embedding‑based candidate generation with LLM‑driven selection, enabling precise and recall‑friendly alignment of ICD versions.
    
    \item Our proposed method mitigate the precision–recall–coverage trade‑off inherent in selection-based methods, delivering high precision with comparable recall and mapping coverage.

    \item Empirically, we demonstrate balanced performance on chapter-wise mapping across different ICD version pairs (i.e., ICD-9-CM$\leftrightarrow$ICD-10-CM and ICD-10-AM$\leftrightarrow$ICD-11) and three disease chapters: across three disease chapters: Diseases of the Digestive System (\textbf{Dig}), Infectious and Parasitic Diseases (\textbf{Inf}) and Diseases of the Respiratory System (\textbf{Resp}.
\end{enumerate}

\section{Preliminaries}
\subsection{International Classification of Disease (ICD)}
\begin{figure}
    \centering
    \includegraphics[width=\linewidth]{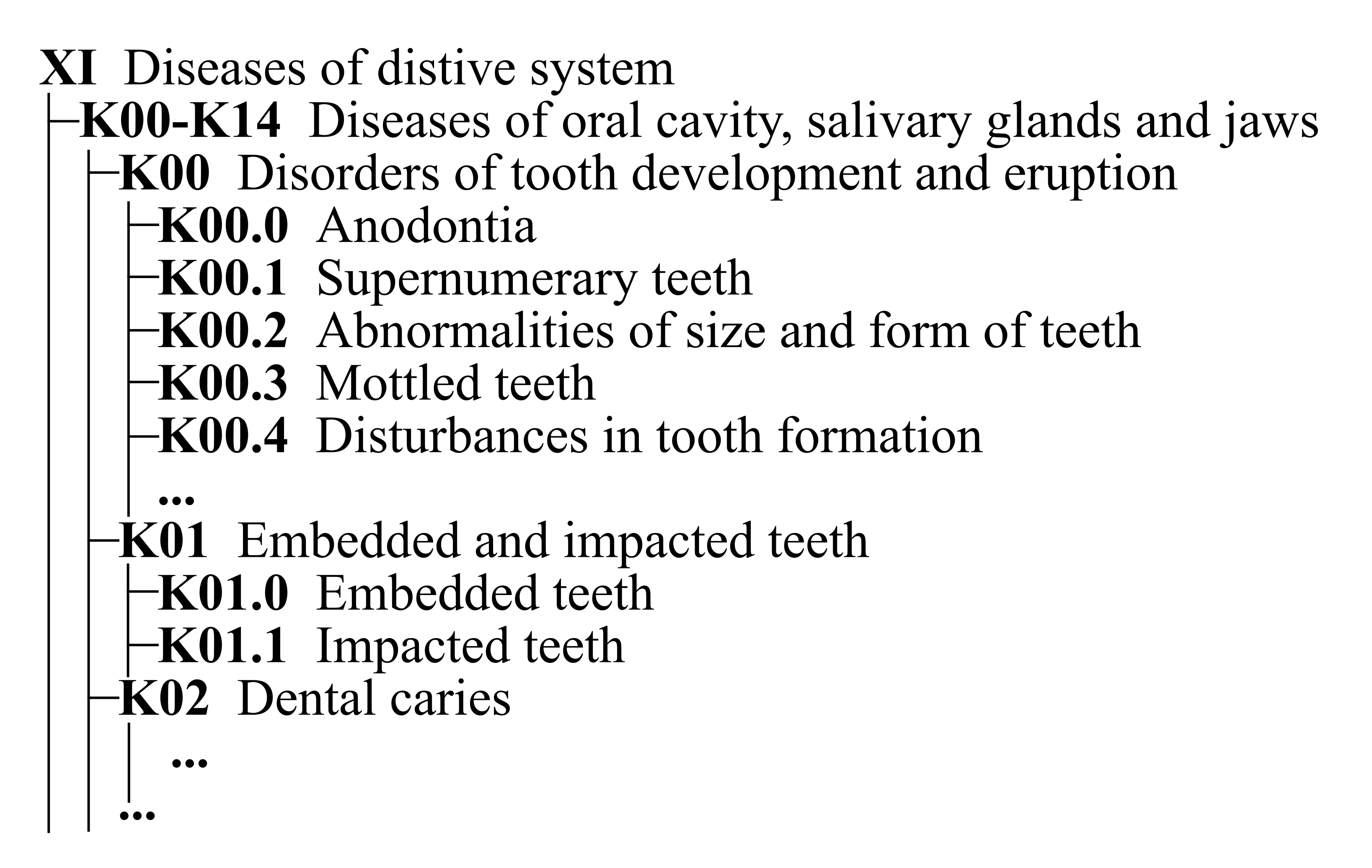}
    \caption{Example of ICD-10's hierarchical structure. The roman numerals represent the chapters (e.g. \textbf{XI}--\emph{Diseases of the digestive system}), while the alphanumeric range represents block (e.g., \textbf{K00-K14}--\emph{Diseases of oral cavity, salivary glands and jaws}). Similarly, a three-character code represents a specific categorie, such as \textbf{K00}--\emph{Disorders of tooth development and eruption}, and subcategories with decimal extensions, like \textbf{K00.1}--\emph{Supernumerary teeth}.}
    \label{fig:icd-10}
    \Description{Illustration of ICD-10's hierarchical organization of diseases.}
\end{figure}
ICD is a hierarchical system that organizes diseases into different chapters, blocks, and groups based on anatomical site, etiology and severity. As shown in Figure~\ref{fig:icd-10}, each disease is assigned a unique code (e.g., \textbf{K00.0}) and a corresponding description (e.g., \textit{Anodontia}). Maintained by the World Health Organization (WHO), ICD is frequently revised, for example, from ICD-10 to ICD-11 (the latest version). A key challenge with this transition is the lack of support for forward and backward compatibility. Direct code comparison is not possible, as different versions use different structures. For instance, ICD-9-CM codes are mostly numeric, whereas ICD-10-CM codes are alphanumeric. While comparing code descriptions to find equivalent source-target code pairs offers a simple solution, given the \emph{linguistic variation}~\citep{10.1093/jamia/ocag004, purja2025managing} across ICD versions, it may fail to cover all ICD codes.

\subsection{Task Definition}
\label{subsec:task-defn}
Suppose $\mathcal{S}=\{s_i\}$ and $\mathcal{T}=\{t_j\}$ be the source and target coding systems. Each code $s_i \in \mathcal{S}$ and $t_j \in \mathcal{T}$ is associated with a textual description $d_S(s_i) \in \mathcal{X}$ and $d_T(t_j) \in \mathcal{X}$, respectively, where $\mathcal{X}$ denotes the natural language space. In this work, we define the task of mapping \emph{source-to-target} (i.e. $\mathcal{S}\rightarrow\mathcal{T}$) as generating a mapping set:

\begin{equation}
    \mathcal{M}_{\mathcal{S}\rightarrow\mathcal{T}}=\{(s_i, T_i) \mid s_i \in \mathcal{S},\; T_i \subseteq \mathcal{T}\}
\end{equation}

This formulation naturally accommodates both the \emph{one-to-one} and \emph{one-to-many} associations. If $T_i=\emptyset$, then no valid equivalent maps for $s_i$ exists in $\mathcal{T}$, which is a common scenario when mapping ICD versions that differ significantly in granularity or concept coverage.




\subsection{Generating Maps Using Embeddings}
\label{subsec:maps-emb}
Embedding-based approaches generate the mapping set by first projecting the source and target code descriptions into a shared embedding space. Suppose  For each source code $s_i$, a corresponding set of equivalent target codes $T_i \subseteq \mathcal{T}$ is defined as:

\begin{equation}
    T_i = \{t_j | \mu(f_\theta(d_S(s_i)), f_\theta(d_T(t_j))) \geq \lambda\},
    \label{eq:1}
\end{equation}

where $f_\theta: \mathcal{X}\rightarrow\mathbb{R}^d$ is a pre-trained encoder model (e.g. BERT) with parameters $\theta$ and $d$ is the embedding space dimension. $\mu: \mathbb{R}^d\times\mathbb{R}^d\rightarrow\mathbb{R}$ is the similarity metric (e.g. cosine similarity) and $\lambda \in \mathbb{R}$ is the threshold value. 

Likewise, we can also generate such a mapping set by selecting \emph{top-K} target code, i.e. $T_i = \text{TopK}_{t_j \in \mathcal{T}} \mu(f_\theta(d_S(s_i)), f_\theta(d_T(t_j)))$, where TopK returns the set of top K target codes with the highest cosine similarity scores to the source code $s_i$. 

The threshold-based approach is highly sensitive to the choice of the threshold value $\lambda$. A higher $\lambda$ results in high precision but low recall, whereas a lower $\lambda$ increases recall but reduces precision. Another potential issue with this method is the \emph{mapping coverage}: a higher threshold value results in low coverage, while a lower value in high coverage. Similarly, the \emph{top-K} based approach generally results in low precision but high recall. It is because each source code is aligned exactly to $K$ target codes resulting in $K-1$ false positives for all the \emph{one-to-one} mapping cases, which is very common in ICD mappings.


\subsection{Bidirectional Mapping Expansion (BiMaps)}
\label{sec:bidir-maps}
Alternatively, we can first construct one-to-one mappings in both directions and use them to generate the final mappings. We call this \emph{bidirectional mapping expansion} (BiMaps). Suppose $\mathcal{M}^{1-1}_{\mathcal{S}\rightarrow\mathcal{T}} = \{(s_i, t^*)\}_i$ be the one-to-one mappings from \emph{source-to-target}, where $t^*=\argmax_{t_j \in \mathcal{T}}\mu(f_\theta(d_S(s_i)), f_\theta(d_T(t_j)))$. Likewise, we can construct $\mathcal{M}^{1-1}_{\mathcal{T}\rightarrow\mathcal{S}} = \{(t_j, s^*)\}_j$. Then, for any source code $s_i$, we can define the set of equivalent target codes as:
\begin{equation}
    T_i = \{t_j | (s_i, t_j) \in \mathcal{M}^{1-1}_{\mathcal{S}\rightarrow\mathcal{T}} \vee (t_j, s_i) \in \mathcal{M}^{1-1}_{\mathcal{T}\rightarrow\mathcal{S}}\}
    \label{eq:2}
\end{equation}

The underlying assumption is that if $(s_i, t_j) \in \mathcal{M}^{1-1}_{\mathcal{S} \rightarrow \mathcal{T}}$ and $(t_k, s_i) \in \mathcal{M}^{1-1}_{\mathcal{T} \rightarrow \mathcal{S}}$, then $s_i \rightarrow \{t_j, t_k\}$. In the worst case, each source code will be mapped to one target code with the highest cosine similarity score. Hence, compared to the threshold or top-K based methods, the bidirectional mapping expansion generally yields higher precision, but may result in lower recall, as some valid matches might not be top-ranked in either direction. Given our objective of maximizing precision while maintaining comparable recall and broad mapping coverage, this approach offers a robust alternative to traditional selection methods.

\subsection{Entity Resolution (ER)}
\label{subsec:er}
\emph{Entity resolution} (ER) aims to match records that refer to the same real-world entity~\citep{fellegi1969theory}. The \emph{blocking-and-matching} approach has emerged as a widely adopted pipeline in ER, which consists of two stages. In the blocking stage, a reduced set of record pairs $R \subseteq \mathcal{D} \times \mathcal{D}' = \{(r, r') | r \in \mathcal{D}, r' \in \mathcal{D}'\}$ is generated by filtering out the obvious non-matching pairs, thereby reducing the search space compared to the full space $\mathcal{D}\times\mathcal{D}'$. And in the matching stage, the candidate pairs in $R$ are evaluated to identify the true matches, which is typically formulated as an independent binary classification task:

\begin{equation}
    \Phi:R \rightarrow \{0, 1\}, (r, r') = \left\{
    \begin{array}{ll}
        1, & \text{if } \phi(r) = \phi(r') \\
        0, & \text{Otherwise}
    \end{array}
    \right.
    \label{eq:3}
\end{equation}

where $\Phi$ is the matching method, and $\phi:\mathcal{D}\cup \mathcal{D}' \rightarrow \mathcal{E}$ and $\mathcal{E}$ denotes set of unique real-world entities. 

\section{Mapping by Blocking and Matching}
\label{sec:method}

Although the goal of ER is significantly different than that of mapping between ICD versions, the \emph{blocking-and-matching} framework is still applicable in the later case. In ER, once we obtain a matching set $M_i \subseteq \mathcal{S} \times \mathcal{T} = \{(s_i, t_j) | s_i \in \mathcal{S}, t_j \in \mathcal{T}\}$, the usual practice is to create clusters of the records that refer to the same real-world entity by taking the transitive closure of $M_i$~\citep{benjelloun2009swoosh}.

In contrast, in ICD mapping we are not interested in forming transitive clusters. Instead, we are interested in identifying all the target codes associated with each source codes. Following our definition of ICD mapping in Section \ref{subsec:task-defn}, the mapping task can be reformulated as:

\begin{equation}
    \mathcal{M}'_{\mathcal{S}\rightarrow\mathcal{T}} = \{(s_i, T_i) | s_i \in \mathcal{S}, T_i=\{t_j|(s_i, t_j) \in R_i\}\}
\end{equation}

Figure~\ref{fig:mapping-pipeline} illustrates our proposed \emph{blocking-and-matching} framework for automatic ICD mapping. The following sub-sections present details of each stage.

\subsection{Blocking Stage} 
\label{sub-sec:block}

There is an inherent trade-off between \emph{effectiveness} and \emph{efficiency} when designing blocking strategies~\cite{papadakis2020blocking}. Effectiveness refers to the blocking method's ability to retain true matching pairs within the generated candidate sets; a robust blocking method must maximize recall. Efficiency, on the other hand, measures the reduction in computational effort achieved by minimizing the number of candidate pairs that require detailed comparison in the matching stage. An ideal blocking strategy must strike a balance between these competing objectives: \textit{reducing the search space sufficiently to make matching tractable, while preserving as many valid pairs as possible.}

The importance of minimizing block size is further emphasized by our matching approach, which utilizes an LLM as a selector through multiple-choice question (MCQ) prompting. Recent studies have demonstrated the detrimental impact of context length on LLM performance. For instance, \citet{liu2024lost} identified the ``\emph{lost in the middle}'' phenomenon, where LLMs struggle to utilize information embedded in longer contexts effectively. Likewise, \citet{du2025context} showed that context length alone can degrade performance, even with perfect retrieval.

These findings are critical for our MCQ prompting strategy: presenting an LLM with 100 candidates may reduce matching accuracy compared to presenting 10 carefully selected candidates, even if the former achieves higher blocking recall. The larger candidate set increases context length and introduces distraction, affecting the model's ability to identify the correct match. Therefore, our blocking strategy must achieve high recall while minimizing the size of the candidate set.

Our blocking method employs a hybrid strategy that combines the top-K and bidirectional mapping expansion methods (Top-$K$+BiMaps) to construct the blocks of candidate sets. Suppose for any $s_i \in \mathcal{S}$, let $T_i^k \subseteq \mathcal{T} \text{ and } T_i^b \subseteq \mathcal{T}$ be the sets of potential matches obtained using the top-K and bidirectional mapping expansion methods respectively. We then generate a block of source and target code pairs  $R_i$ as:
\begin{equation}
    R_i \subseteq \mathcal{S} \times \mathcal{T} = \{(s_i, t_j) | t_j \in T_i^k \cup T_i^b\}
\end{equation}

Empirically, we found that the top-$K$ method achieved similar recall to the threshold-based method but with significantly fewer target codes per source code (see Appendix~\ref{app:card_block-size-comp}), making it more computationally efficient. Hence, we opted for the top-$K$ method as our baseline blocking strategy.

The bidirectional mapping expansion method produces candidate sets that largely overlap with top-$K$ selections, differing primarily in edge cases. As a result, the hybrid Top-$K$+BiMaps approach introduces only minimal increases in block size. For a detailed discussion on the comparison between simple top-K selection and our hybrid blocking strategy, see Appendix~\ref{app:card_effect-hybrid-blocking-block-size}.

\subsection{Matching Stage}
\label{sub-sec:match}

\begin{figure}[!ht]
    \begin{lstlisting}[
        basicstyle=\small\ttfamily,
    ]
    You are a clinical terminology expert tasked with identifying semantically equivalent codes.
    
    Source Code Description:
    <SOURCE_CODE_DESC>

    Select ALL options that represent the same or highly similar clinical condition as the source.
    <CODE_1> <CODE_DESC_1>
    <CODE_2> <CODE_DESC_2>
    ...
    
    Instructions:
    Return a **single Python-style list**:
    - Include only the matching code values.
    - If none match, return '[]'.
    - **No explanation. No extra text. Only the list.**
    \end{lstlisting}
    \caption{Prompt template for our LLM-based selector framework. We use the source code description as the anchor term and provide a list of target codes as potential matches, instructing the LLM to select all options that are semantically equivalent to the anchor. \texttt{<SOURCE\_CODE\_DESC>} is the placeholder for the source code description, while \texttt{<CODE\_*>} and \texttt{<CODE\_DESC\_*>} are placeholders for target codes and their descriptions, respectively.}
    \label{fig:prompt-template}
    \Description{Prompt template used for LLM-based selector framework.}
\end{figure}

In the blocking phase, rather than generating the record pair sets $R_i \subseteq \mathcal{S} \times \mathcal{T}$, we obtain a set of candidate target codes for each source code (i.e. $R'_i \subseteq \mathcal{T} = \{t_j| (s_i, t_j) \in R_i\}$). Following \cite{wang2024match}, the matching task is then formulated as selecting the best matching target code(s) from $R'_i$ for a source code $s_i$, as opposed to performing the pair-wise binary classification for all $(s_i, t_j) \in R$ pairs as in Eq. \ref{eq:3}.

Following recent progress in \emph{zero-shot} and \emph{few-shot} large language model (LLM)-based matching approaches~\citep{narayan2022can, wang2024match, xu2024kcmf}, we employ an LLM to select matching options. Specifically, we construct prompts using a predefined template and ask the LLM to identify the most appropriate matches. Unlike existing methods in entity resolution (ER), which explicitly instruct the LLM to select a single matching option, our approach allows the LLM to choose one or more matching options:

\begin{equation}
    \label{eq:matching-llm}
    \text{LLM}: \tau(s_i, R'_i) \rightarrow \{0, 1\}^{|R'_i|},
\end{equation}
where \( \tau(s_i, R'_i) \) is a template function that takes a source code \( s_i \) and a set of candidate target codes \( R'_i \), and returns a multiple-choice question (MCQ) prompt suitable for querying the LLM (see Figure~\ref{fig:prompt-template} for the prompt template used in this study). We provide an example of the input prompt and the LLM response in Appendix~\ref{app:prompt-eg}.


In this work, we use \emph{Qwen3-8B}~\citep{yang2025qwen3} as our preferred LLM, as it provides an explicit mechanism to enable ( or disable) intermediate reasoning steps. While in our previous work~\citep{10.1093/jamia/ocag004}, we showed that the prompting framework offers a robust solution for ICD mappings (in particular, handling granular inconsistencies across ICD versions), more effective prompt engineering, such as chain-of-thought (CoT)~\citep{sinha2025dr} and few-shot prompting~\citep{brown2020language}, could further improve the matching performance; we leave this exploration to future work.

\section{Experiment Details}
\subsection{Dataset}
For evaluation, we performed mappings between different ICD versions, namely ICD-9-CM$\leftrightarrow$ICD-10-CM and ICD-10-AM$\leftrightarrow$ICD-11, across three disease chapters: Disease of Digestive System (\textbf{Dig}), Infectious and Parasitic Diseases (\textbf{Inf}), and Disease of Respiratory System (\textbf{Resp}). For ICD-9-CM$\leftrightarrow$ICD-10-CM, we used the General Equivalence Maps (GEMs)~\footnote{\url{https://www.cms.gov/medicare/coding-billing/icd-10-codes/icd-10-cm-icd-10-pcs-gem-archive}} as the ground truth. For ICD-10-AM$\leftrightarrow$ICD-11, we employed a sequential mapping approach similar to~\citep{xu2022sequential}, using ICD-10 as the reference version (i.e., ICD-10-AM$\leftrightarrow$ICD-10$\leftrightarrow$ICD-11. For ICD-10-AM$\leftrightarrow$ICD-10, we used the mapping tables provided by Independent Health and Aged Care Pricing Authority (IHACPA)~\footnote{\url{https://www.ihacpa.gov.au/resources/icd-10-am-and-achi-mapping-tables}}. And for ICD-10$\leftrightarrow$ICD-11, we used the mapping tables provided by the World Health Organization (WHO)~\footnote{\url{https://icd.who.int/browse/2025-01/mms/en}}. Further details on the datasets are provided in Appendix~\ref{app:data-details}.


\subsection{Obtaining Embeddings}
Obtaining better embeddings for the ICD code descriptions is crucial for generating highly relevant blocks of target codes. In this work, we adopt the data-augmentation techniques introduced in one of our previous work~\citep{10.1093/jamia/ocag004} to obtain embeddings for ICD code descriptions. Specifically, we first generate the \emph{hierarchy-augmented} and \emph{LLM-generated} descriptions for each source code, which are then encode using a pre-trained encoder model and compute their mean to form the final representation. We use \emph{sentence-transformer} (SBERT)~\cite{reimers-2019-sentence-bert} as the encoder model and Qwen3-8B to generate clinical descriptions.

\subsection{Baseline}
We compare our proposed methods against two standard selection strategies: threshold-based selection with $\lambda=0.85$ and top-$K$ selection with $K=5$. For our hybrid blocking strategy (Top-$K$+BiMaps), we set $K=5$. The threshold value of 0.85 provides a more balanced trade-off between precision, recall, and coverage (see Appendix~\ref{app:precision-recall-tradeoff} for a detailed comparison).

\subsection{Evaluation Metrics}
For evaluation, we report \emph{Precision}, \emph{Recall}, and \emph{F1-score} using both macro and micro averaging. Macro scores are computed by averaging the corresponding metric across all classes, while micro scores are calculated by aggregating the counts of true positives, false positives, and false negatives. For example, the macro and micro precisions are defined as:
\begin{equation}
    \text{Macro-Precision} = \frac{1}{N}\sum_i\frac{TP_i}{TP_i+FP_i},
\end{equation}
\begin{equation}
    \text{Micro-Precision} = \frac{\sum_i TP_i}{\sum_i TP_i + FP_i},
\end{equation}
where $TP_i \text{ and } FP_i$ are the true positives and false positives for $i^{th}$ source code.

We compute the F1-scores by taking the harmonic mean of corresponding precision and recall. As demonstrated in our previous work~\citep{10.1093/jamia/ocag004} using the same model on similar multiple-choice prompts, LLM outputs showed minimal variance across independent runs; we therefore report single-inference results.

Likewise, we also report the \emph{mapping coverage} ($MC$), which is defined as the proportion of source codes that is mapped to at-least one target code.
\begin{equation}
    MC = \frac{|\{j | (s_j, T_j) \in \mathcal{M}_{\mathcal{S}\rightarrow\mathcal{T}} \text{ and } T_j \neq \emptyset \}|}{N}
\end{equation}

\section{Results and Discussion}
\subsection{Main Results}

We present our main results in Table~\ref{tab:card_main_results}, comparing our proposed methods, including the bidirectional mapping expansion and blocking-and-matching framework, against the chosen baseline methods across different ICD mapping pairs and chapters. Macro-averaged results are provided in Appendix~\ref{app:card_macro-avg-results} and show consistent patterns with the micro-averaged metrics reported here.

\begin{table*}[!ht]
    \centering
    \caption{Performance comparison of baseline selection methods, bidirectional mapping expansion, and the blocking-and-matching pipeline across four mapping tasks and three disease chapters: Diseases of the Digestive System (\textbf{Dig}), Infectious and Parasitic Diseases (\textbf{Inf}), and Diseases of the Respiratory System (\textbf{Resp}). The reported numbers are micro-averaged precision (\textbf{Pr}), recall (\textbf{Rc}), and F1-score (\textbf{F1}). \textbf{MC} represents mapping coverage (the proportion of source codes receiving at least one target mapping).}
    
    \begin{subtable}[t]{0.48\linewidth}
        \centering
        \resizebox{\linewidth}{!}{
            \begin{tabular}{lcccccccccccc}
                \toprule
                & \multicolumn{4}{c}{\textbf{Dig}} & \multicolumn{4}{c}{\textbf{Inf}} & \multicolumn{4}{c}{\textbf{Resp}} \\
                \cmidrule(lr){2-5} \cmidrule(lr){6-9} \cmidrule(lr){10-13}
                & \textbf{Pr} & \textbf{Rc} & \textbf{F1} & \textbf{MC} & \textbf{Pr} & \textbf{Rc} & \textbf{F1} & \textbf{MC} & \textbf{Pr} & \textbf{Rc} & \textbf{F1} & \textbf{MC} \\
                \midrule
                Threshold ($\lambda=0.85$) & 0.19 & 0.82 & 0.31 & 0.93 & 0.41 & 0.63 & 0.50 & 0.73 & 0.39 & 0.66 & 0.49 & 0.83\\
                Top-$K$ ($K=5$) & 0.21 & \textbf{0.90} & 0.34 & \underline{1.00} & 0.20 & \textbf{0.88} & 0.33 & \underline{1.00} & 0.20 & \textbf{0.83} & 0.32 & \underline{1.00} \\
                \midrule
                BiMaps & 0.58 & 0.85 & 0.69 & \underline{1.00} & 0.55 & 0.79 & 0.65 & \underline{1.00} & 0.55 & 0.73 & 0.63 & \underline{1.00} \\
                \midrule
                Blocking-and-Matching & \textbf{0.71} & 0.85 & \textbf{0.77} & \underline{1.00} & \textbf{0.63} & 0.81 & \textbf{0.71} & \underline{1.00} & \textbf{0.65} & 0.75 & \underline{0.70} & \underline{1.00} \\
                \bottomrule
            \end{tabular}
        }
        \caption{ICD-9-CM $\rightarrow$ ICD-10-CM}
        \label{tab:card_main-results-a}
    \end{subtable}\hfill
    \begin{subtable}[t]{0.48\linewidth}
        \centering
        \resizebox{\linewidth}{!}{
            \begin{tabular}{lcccccccccccc}
                \toprule
                & \multicolumn{4}{c}{\textbf{Dig}} & \multicolumn{4}{c}{\textbf{Inf}} & \multicolumn{4}{c}{\textbf{Resp}} \\
                \cmidrule(lr){2-5} \cmidrule(lr){6-9} \cmidrule(lr){10-13}
                & \textbf{Pr} & \textbf{Rc} & \textbf{F1} & \textbf{MC} & \textbf{Pr} & \textbf{Rc} & \textbf{F1} & \textbf{MC} & \textbf{Pr} & \textbf{Rc} & \textbf{F1} & \textbf{MC} \\
                \midrule
                Threshold ($\lambda=0.85$) & 0.16 & 0.71 & 0.26 & 0.78 & 0.52 & 0.60 & 0.56 & 0.68 & 0.39 & 0.56 & 0.46 & 0.71\\
                Top-$K$ ($K=5$) & 0.20 & \textbf{0.87} & 0.33 & \underline{1.00} & 0.20 & \textbf{0.91} & 0.33 & \underline{1.00} & 0.19 & \textbf{0.80} & 0.31 & \underline{1.00}\\
                \midrule
                BiMaps & 0.55 & 0.73 & 0.63 & \underline{1.00} & \underline{0.70} & 0.78 & 0.74 & \underline{1.00} & 0.57 & 0.64 & 0.60 & \underline{1.00} \\
                \midrule
                Blocking-and-Matching & \textbf{0.64} & 0.80 & \textbf{0.71} & 0.99 & \underline{0.70} & 0.83 & \textbf{0.76} & 0.99 & \textbf{0.60} & 0.68 & \textbf{0.64} & 0.98 \\
                \bottomrule
            \end{tabular}
        }
        \caption{ICD-10-CM $\rightarrow$ ICD-9-CM}
        \label{tab:card_main-results-b}
    \end{subtable}
    
    \vspace{0.01\linewidth}
    
    \begin{subtable}[t]{0.48\linewidth}
        \centering
        \resizebox{\linewidth}{!}{
            \begin{tabular}{lcccccccccccc}
                \toprule
                & \multicolumn{4}{c}{\textbf{Dig}} & \multicolumn{4}{c}{\textbf{Inf}} & \multicolumn{4}{c}{\textbf{Resp}} \\
                \cmidrule(lr){2-5} \cmidrule(lr){6-9} \cmidrule(lr){10-13}
                & \textbf{Pr} & \textbf{Rc} & \textbf{F1} & \textbf{MC} & \textbf{Pr} & \textbf{Rc} & \textbf{F1} & \textbf{MC} & \textbf{Pr} & \textbf{Rc} & \textbf{F1} & \textbf{MC} \\
                \midrule
                Threshold ($\lambda=0.85$) & 0.42 & 0.58 & 0.49 & 0.82 & 0.40 & 0.67 & 0.50 & 0.81 & 0.39 & 0.75 & 0.51 & 0.87\\
                Top-$K$ ($K=5$) & 0.19 & \textbf{0.74} & 0.30 & \underline{1.00} & 0.19 & \textbf{0.87} & 0.31 & \underline{1.00} & 0.19 & \textbf{0.89} & 0.31 & \underline{1.00}\\
                \midrule
                BiMaps & 0.39 & 0.68 & 0.50 & \underline{1.00} & 0.51 & 0.69 & 0.59 & \underline{1.00} & 0.54 & 0.75 & 0.63 & \underline{1.00} \\
                \midrule
                Blocking-and-Matching & \textbf{0.47} & 0.64 & \textbf{0.54} & \underline{1.00} & \textbf{0.55} & 0.74 & \textbf{0.63} & \underline{1.00} & \textbf{0.58} & 0.76 & \textbf{0.66} & \underline{1.00} \\
                \bottomrule
            \end{tabular}
        }
        \caption{ICD-10-AM $\rightarrow$ ICD-11}
        \label{tab:card_main-results-c}
    \end{subtable}\hfill
    \begin{subtable}[t]{0.48\linewidth}
        \centering
        \resizebox{\linewidth}{!}{
            \begin{tabular}{lcccccccccccc}
                \toprule
                & \multicolumn{4}{c}{\textbf{Dig}} & \multicolumn{4}{c}{\textbf{Inf}} & \multicolumn{4}{c}{\textbf{Resp}} \\
                \cmidrule(lr){2-5} \cmidrule(lr){6-9} \cmidrule(lr){10-13}
                & \textbf{Pr} & \textbf{Rc} & \textbf{F1} & \textbf{MC} & \textbf{Pr} & \textbf{Rc} & \textbf{F1} & \textbf{MC} & \textbf{Pr} & \textbf{Rc} & \textbf{F1} & \textbf{MC} \\
                \midrule
                Threshold ($\lambda=0.85$) & 0.37 & 0.49 & 0.42 & 0.57 & 0.43 & 0.69 & 0.53 & 0.77 & 0.47 & 0.69 & 0.56 & 0.72\\
                Top-$K$ ($K=5$) & 0.19 & \textbf{0.71} & 0.30 & \underline{1.00} & 0.19 & \textbf{0.8}8 & 0.31 & \underline{1.00} & 0.17 & \textbf{0.84} & 0.28 & \underline{1.00} \\
                \midrule
                BiMaps & \textbf{0.59} & 0.53 & \textbf{0.56} & \underline{1.00} & 0.60 & 0.72 & 0.65 & \underline{1.00} & 0.64 & 0.74 & 0.69 & \underline{1.00} \\
                \midrule
                Blocking-and-Matching & 0.54 & 0.53 & 0.53 & 0.99 & \textbf{0.59} & 0.76 & \textbf{0.66} & 0.99 & \textbf{0.63} & 0.77 & \textbf{0.69} & 0.99 \\
                \bottomrule
            \end{tabular}
        }
        \caption{ICD-11 $\rightarrow$ ICD-10-AM}
        \label{tab:card_main-results-d}
    \end{subtable}
    \label{tab:card_main_results}
\end{table*}

\paragraph{Comparison of Selection Methods: Bidirectional Mapping Expansion vs. Baseline Methods.} The baseline selection methods demonstrate a fundamental trade-off between precision and recall. The threshold-based approach ($\lambda=0.85$) achieved moderate precision, ranging from 0.16 to 0.52, but exhibits highly variable recall, which ranges from 0.49 to 0.82. More critically, mapping coverage is inconsistent and often inadequate, with values as low as 0.57 for the ICD-11 $\rightarrow$ ICD-10-AM within the Disease of the Digestive System chapter. This indicates that the fixed similarity threshold fails to identify any candidate mappings for a substantial proportion of source codes, limiting the method's practical utility.

The top-$K$ method ($K=5$) addressed the coverage problem by guaranteeing that every source code receives exactly $K$ candidate mappings, achieving perfect mapping coverage (MC = 1.00) across all tasks. However, this comes at a severe cost to precision, which dropped to 0.17--0.21 in most cases. While the method achieved the highest recall among all approaches (0.71--0.91), the indiscriminate selection of $K$ candidates regardless of semantic similarity results in the inclusion of numerous false positives. Therefore, the resulting F1-scores (0.28--0.34) are consistently lower than those for the threshold-based approach (0.26--0.56), despite having superior recall.

Bidirectional mapping expansion (BiMaps) represented a substantial improvement over both baseline methods. It achieved superior precision compared to the chosen baselines, while maintaining a perfect mapping coverage (MC = 1.00) across all tasks and chapters. BiMaps improved precision by an average of 0.19 over threshold-based selection and 0.37 over the top-$K$ baseline. On certain tasks, the method demonstrated strong performance, such as the ICD-10-CM $\rightarrow$ ICD-9-CM for Infectious and Parasitic Diseases chapter, where it achieved a precision of 0.70 and an F1 score of 0.74.

However, BiMaps exhibited recall scores (0.53--0.85) that generally remained below the levels achieved by the Top-K baseline (0.71--0.91). Regardless, this trade-off aligns with our objective of optimizing precision while maintaining comparable recall and mapping coverage. BiMaps successfully achieved this balance, delivering F1-scores (0.50--0.74) that consistently outperformed both baseline methods, while ensuring complete mapping coverage (MC = 1.00) across all tasks and chapters.

\paragraph{Precision Improvements with Blocking-and-Matching Pipeline.} Our proposed blocking-and-matching framework achieved superior performance by combining the candidate generation strengths of Top-K and BiMaps with an LLM-based matching approach. The framework consistently achieved the highest precision across nearly all mapping pairs and chapters, with values ranging from 0.47 to 0.71; this represented average precision improvements of 2-3.5 times over the Top-K baseline, and gains of 5--35\% points over BiMaps in most mapping tasks (except for ICD-11 $\rightarrow$ ICD-10-AM for Diseases of the Digestive System, where BiMaps yielded better precision of 0.59 compared to 0.54 for the blocking and matching approach).

Most importantly, these improvements in precision were accomplished without compromising recall. The framework maintained recall levels (0.53--0.85) that were competitive with or superior to BiMaps and comparable to the high-recall Top-K baseline, despite operating with substantially higher precision. For example, on the ICD-9-CM $\rightarrow$ ICD-10-CM task for Disease of the Digestive System chapter, the blocking-and-matching approach achieved 0.71 precision and 0.85 recall, compared to BiMaps' 0.58 precision and 0.85 recall. The improved balance between precision and recall translated directly into superior F1-scores. The blocking-and-matching framework achieved the highest F1-score in 11 out of 12 mapping tasks, with values ranging from 0.53 to 0.77. 

The framework also maintained near-perfect mapping coverage (MC $\geq$ 0.98) across all configurations, ensuring that almost every source code received at least one candidate mapping. This addressed the critical limitation of threshold-based approaches while avoiding the precision penalties of indiscriminate Top-K selection.


\subsection{Discussion}
Our results demonstrated that the blocking-and-matching framework achieved substantial improvements in automated ICD code mapping by effectively balancing precision, recall, and coverage. We discuss the key findings and their implications for practical deployment.

\paragraph{Limitations of Simple Selection Strategies.} The baseline methods revealed fundamental trade-offs that limited their practical utility. The threshold-based approach's inconsistent mapping coverage, as low as 0.57 for some chapters, represents a critical failure mode, as it reduces the usability of the generated maps. Conversely, the top-$K$ method achieved perfect coverage but with precision dropping to 0.17--0.21, thereby reducing the reliability of the generated maps.

These results underscore that maximizing a single metric is insufficient; instead, a comprehensive approach is necessary that must simultaneously achieve adequate precision, recall, and mapping coverage, requirements that neither baseline satisfied.


\paragraph{Bidirectional Consistency Improved Precision While Maintaining Comparable Recall and Complete Mapping Coverage.} BiMaps use both forward and backward $1:1$ mappings between the source and target ICD versions, yielding high-quality, semantically similar maps. This results in substantial improvements in precision over Top-K selection (averaging 0.37), while ensuring a complete mapping coverage.

The lower recall compared to Top-K (0.53--0.85 vs. 0.71--0.91) reflects an inherent limitation of the BiMaps: semantically equivalent ICD code descriptions do not always exhibit the highest similarity in the embedding space. This suggests that embedding-based similarity alone, even when applied bidirectionally, cannot fully capture the nuanced semantic relationships required for code mapping.

Nevertheless, the superior F1-scores (0.50--0.74) compared to both baselines demonstrated that BiMaps achieved the desired balance for our objectives. The precision gains outweighed the recall penalties, indicating that BiMaps removed more false positives than true positives, which can be a more favourable outcome for the manual review process.

\paragraph{LLM-Based ICD Code Mapping: Performance Superiority and Computational Trade-offs.} The blocking-and-matching framework achieved higher precision over both top-$K$ (2--3.5 times) and BiMaps (5--35 percentage points) without sacrificing recall, demonstrating the effectiveness of training-free, zero-shot prompting for code mapping. The LLM's ability to identify valid mappings without explicit training on the mapping task suggests that general-purpose medical knowledge acquired during pre-training can be successfully transferred to this specialized application. This training-free approach offers significant practical advantages: the framework remains flexible and adaptable. It can readily incorporate newer, larger models trained on more extensive corpora with greater compute resources, potentially further enhancing mapping performance without requiring task-specific fine-tuning or labeled mapping data.

The LLM-based approach introduces substantial computational requirements. For instance, for our default configuration ($K=5$ and \verb|batch_size=8|), the blocking-and-matching framework took 43 hours, 38 minutes, and 52 seconds to process the mappings between the ICD pairs across three chapters (see Appendix~\ref{app:card_compute-time} for details on compute time). Extending this approach to map entire ICD versions could result in prohibitively long processing times. However, given that manual mapping, starting from scratch, is often very time-consuming (e.g., WHO's ICD-10$\rightarrow$ICD-11 is still under active development) and typically involves a large number of trained professionals, requiring billions of dollars in funding, the computational overhead is justifiable. 

Furthermore, major ICD version transitions occur infrequently, often spanning decades between releases, and revisions within the same version result in minor changes. Hence, it represents an upfront one-time investment during major version transitions; once the maps are created, they are reused for the downstream tasks.

\paragraph{Sensitivity Analysis: The Precision-Recall Trade-off in Blocking}
Theoretically, increasing the block size would provide the LLM with a broader set of candidates, thereby improving recall and overall mapping performance. However, our ablation study (see Appendix~\ref{app:ablation}) demonstrates diminishing returns from this strategy. Specifically, we observed that as the block size $K$ increases, precision decreases significantly, while the gain in recall remains marginal. This phenomenon occurs because the additional candidates retrieved at higher values of $K$ often lie close to one another in the shared embedding space due to shared clinical context (e.g., related symptoms or anatomical sites). However, they do not represent semantically equivalent mappings. By increasing the block size, we introduce a higher density of these 'near-miss' distractors into the MCQ prompt, which can mislead the LLM into selecting related but strictly invalid terms. 

\paragraph{Performance Variations Across Mapping Pairs and Chapters.} Performance varied across mapping pairs and chapters. The ICD-9-CM $\leftrightarrow$ ICD-10-CM mapping pairs yielded better precision and recall compared to ICD-10-AM $\leftrightarrow$ ICD-11, with the blocking-and-matching framework achieving 16.96\% higher precision and 11.43\% higher recall on average across the three chapters. These differences could be linked to the reliability of the ground-truth mappings between ICD-10-AM and ICD-11. Since we used sequential mapping to automatically generate these ground-truth mappings using ICD-10 as the reference version, expert evaluation would be needed to assess the true performance on these tasks. Nevertheless, these results suggest that mapping between an established version and an emerging version (e.g., ICD-10-AM $\leftrightarrow$ ICD-11) presents different challenges compared to mapping between well-established versions (e.g., ICD-9-CM $\leftrightarrow$ ICD-10-CM).

Similarly, the Diseases of the Digestive System chapter exhibited the highest variability in both precision (0.47--0.71) and recall (0.53--0.85), suggesting that mapping difficulty in this domain is highly dependent on the specific ICD version pair. In contrast, Infectious and Parasitic Diseases demonstrated more consistent performance, with precision ranging from 0.55 to 0.70 and recall ranging from 0.74 to 0.81, indicating more standardized terminology and clearer categorical boundaries in this domain. Respiratory System diseases showed intermediate variability, with precision ranging from 0.58 to 0.65 and recall from 0.68 to 0.77. These differences likely reflect varying levels of diagnostic complexity and terminological standardization across medical specialties. The observed variability further suggests that rapidly evolving and structurally complex chapters, such as Neoplasms, may present even greater mapping challenges due to their evolving classification systems and increased coding complexity.

\paragraph{Granularity Mismatches and False Positives.} A notable source of false positives in our evaluation stems from granularity mismatches between predicted mappings and ground-truth assignments. Our analysis revealed that when the blocking-and-matching framework predicted parent target codes, 11.30\%--36.57\% of these predictions were hierarchically valid parents of the ground-truth child code rather than genuinely incorrect mappings (see Appendix~\ref{app:card_granularity_mismatch} for detailed breakdown by ICD version pair and chapter). This limitation suggests that the reported precision values may underestimate the true performance of the blocking-and-matching framework.

\section{Related Work}
\emph{Entity resolution} (ER) has been a long-standing challenge in data integration \citep{fellegi1969theory, sehgal2006entity, gal2014uncertain, brunner2020entity}. A core methodology in ER  is the \emph{blocking-and-matching} pipeline, where the \emph{blocking} stage significantly reduces the candidate pairs by filtering out the obvious dissimilar records, and the \emph{matching} stage identifies matching record pairs using various techniques.

Traditional approaches to matching include the rule-based matching method, which applies domain-specific knowledge to construct a hierarchical set of matching rules to find a match \citep{agoun2022access,zhang2020similar}, and probabilistic methods estimate the likelihood of a match based on agreement patterns across fields/attributes \citep{xu2022simple, moretti2023improving, fellegi1969theory}. Deep learning-based methods cast the matching problem as an independent pair-wise binary classification task. The advent of Large Language Models (LLMs) has introduced a zero- or few-shot paradigm to entity matching \citep{narayan2022can, xu2024kcmf, wang2024match}. Most of these methods interface an LLM with an appropriate prompt to classify a pair of records as a match (or no match), except \cite{wang2024match}, which uses the LLM as \emph{matcher, comparator} and \emph{selector} to find a match.

We draw inspiration from this robust ER framework for our ICD mapping task. However, a critical distinction in our problem, unlike the typical \emph{one-to-one} (each record matched to at most one record) assumption in the traditional ER, is the existence of \emph{one-to-many} associations in ICD mappings. Our approach, specifically, accounts for this cardinality characteristic, extending beyond a simple binary classification to identify and accommodate all relevant target codes for a given source code.

\section{Conclusion}
In this work, we addressed a practical challenge in the automatic mapping of disease classification systems, such as ICD, posed by embedding-based selection methods when managing mapping cardinality: the trade-offs among precision, recall, and mapping coverage. High precision but low recall and mapping coverage would leave a large gap for the expert validators, requiring them to generate maps from scratch in most cases. On the other hand, high recall with complete mapping coverage but low precision would significantly increase the workload for validators, as it would include many false positives. Our proposed \emph{blocking-and-matching} framework addresses this trade-off by yielding balanced performance, i.e., high precision with comparable recall and broader mapping coverage, thereby helping to accelerate the manual validation process.

In the context of health informatics, this efficiency is critical for downstream applications. Access to reliable mappings enables more accurate epidemiological surveillance, and tracking disease patterns across populations and time periods. They support better health services research by enabling more accurate analysis of treatment outcomes and healthcare utilization. They facilitate more robust public health policy evaluation by ensuring consistent measurement of health outcomes despite changes in classification systems.

\section{Limitations and Ethical Considerations}
\paragraph{Limitations.} While the blocking-and-matching framework produced balanced results compared to traditional selection approaches, several limitations require considerations. First, our evaluation covered mappings between different country-specific versions and the international version across three disease chapters. Given the performance variation across different version pairs and chapters, future work should investigate how the performance gains translate to other mapping scenarios, such as cross-country version mappings (e.g., ICD-9-CM $\leftrightarrow$ ICD-10-AM) and other chapters such as Neoplasms.

Second, all ICD versions in our study were in English; however, many country-specific versions use different languages (e.g., ICD-10-GM in German). Extending the blocking-and-matching pipeline to mappings between ICD versions in heterogeneous languages would require: (a) embedding models capable of projecting code descriptions from different languages into a shared semantic space where proximity reflects semantic similarity, and (b) LLMs capable of processing multilingual prompts for the matching stage. Recent advances in multilingual embeddings and large language models~\citep{huang2024survey} suggest these extensions are feasible, though empirical validation across language pairs would be necessary.

Third, given granular inconsistencies across different ICD versions (source-to-parent mappings), it is crucial to address them to provide a more comprehensive end-to-end mapping solution. The \emph{residual code}\footnote{The residual codes are categories used to classify clinical condition that is either non-specific or does not fit into a more defined, explicit category, such as ``unspecified'' and ``other specified''.} selection offers a simple solution; however, in our previous work~\citep{10.1093/jamia/ocag004}, we showed that it can be highly restrictive, particularly when a parent code lacks such residual codes, resulting in the exclusion of correct parent-level maps. Instead, we can select either all child codes or a few with the highest cosine similarity scores as potential matches, and include them as options in the prompt. However, this may significantly increase the block size, particularly when multiple such options exist. This represents a promising direction for future work to refine mapping quality while managing computational constraints.

\paragraph{Ethical Considerations.} This research exclusively utilizes publicly available, non-identifiable administrative data comprising disease classification codes and their cross-system mappings. The study does not involve human subjects, personal health records, or identifiable clinical notes. Therefore, it is exempt from human ethics review under Western Sydney University's research ethics policies and the ACM Policy on Research Involving Human Participants.


\section{GenAI Disclosure}
While preparing this manuscript, the author(s) used Grammarly (including Grammarly's generative AI feature) for improving the clarity and correcting grammatical errors in the existing text. Grammarly's generative AI was primarily used to refine phrasing and improve the readability of already-written content. The manuscript does not include any new content, including text, images, tables, code, or other original materials, generated using GenAI tools. The authors take full responsibility for the content and integrity of this work.


\bibliographystyle{ACM-Reference-Format}
\bibliography{bibliography}

\appendix
\section{Dataset Details}
\label{app:data-details}
\subsection{Source}
We used the ICD-9-CM (version 32) from the Centres for Medicare and Medicaid Services (CMS)\footnote{\url{https://www.cms.gov/medicare/coding-billing/icd-10-codes}} and the ICD-10-CM (FY22 release) from the Centers for Disease Control and Prevention (CDC)\footnote{\url{https://www.cdc.gov/nchs/icd/icd-10-cm/files.html}}. Likewise, we used the ICD-10-AM (twelfth edition) provided by the Independent Health and Aged Care Pricing Authority (IHACPA)\footnote{\url{https://www.ihacpa.gov.au/resources/icd-10-amachiacs-twelfth-edition}}. We accessed the ICD-11 codes via the WHO API (version 2.5)\footnote{\url{https://icd.who.int/icdapi}}. It is important to note that the WHO API provides only pre-coordinated ICD-11 codes. Therefore, we used the parent codes in those cases where the source codes are mapped to the post-coordinated codes. 

\subsection{Chapters}
\label{app:data-details:chapters}
We employed a chapter-wise mapping strategy, concentrating on the \emph{Infectious and Parasitic Diseases} (\textbf{Inf}), \emph{Diseases of the Respiratory System} (\textbf{Resp}), and \emph{Diseases of the Digestive System} (\textbf{Dig}) chapters. We used this approach to limit the search space for the potential maps. Also, we include all the three- and four-digit codes. Hence, as shown in Table \ref{tab:chapters}, several codes have no maps because they are either the immediate parents or a broader category in the hierarchy. Additionally, this also include the number of cases where the source codes are mapped completely to different target chapters.

\begin{table}[ht]
	\centering
    \caption{Total number of codes ($N$), and the number of cases where the source code has no maps in ground truth mapping files ($N_{nm}$)}
	\resizebox{\linewidth}{!}{
	\begin{tabular}{lcccccc}
        \toprule
        & \multicolumn{2}{c}{\textbf{Dig}} & \multicolumn{2}{c}{\textbf{Inf}} & \multicolumn{2}{c}{\textbf{Resp}} \\
        \cmidrule(lr){2-3}\cmidrule(lr){4-5}\cmidrule(lr){6-7}
        \textbf{Mapping Direction} & $N$ & $N_{nm}$ & $N$ & $N_{nm}$ & $N$ & $N_{nm}$ \\
        \midrule
        \textbf{ICD-9-CM$\rightarrow$ICD-10-CM} & 757 & 274 & 889 & 0 & 320 & 93 \\
        \textbf{ICD-10-CM$\rightarrow$ICD-9-CM} & 795 & 213 & 1158 & 117 & 369 & 72 \\
        \textbf{ICD-10-AM$\rightarrow$ICD-11} & 617 & 290 & 921 & 207 & 281 & 74 \\
        \textbf{ICD-11$\rightarrow$ICD-10-AM} & 969 & 437 & 1004 & 398 & 342 & 136 \\
        \bottomrule
    \end{tabular}
    }
	\label{tab:chapters}
\end{table}

\section{Ablation Study: Effect of Block Size on Mapping Performance}
\label{app:ablation}
As discussed earlier (Section~\ref{sub-sec:block}), our blocking stage uses a hybrid approach, combining the top-$K$ selection and bidirectional mapping expansion. Therefore, a critical hyperparameter in this process is $K$, which determines the number of candidates selected via the top-$K$ method. While the higher $K$ values yield better recall (\emph{effectiveness}), this may affect the performance of the LLM-based matcher (\emph{efficiency}). Prior studies have demonstrated the detrimental impact of context length on LLM performance~\citep{liu2024lost, du2025context}. To understand how block size affects the LLM-based matching performance, we conducted experiments with $K \in \{5, 7, 10\}$ across all mapping pairs and chapters.

\begin{figure*}[!htb]
    \centering
    \begin{subfigure}{\textwidth}
        \centering
        \includegraphics[width=0.75\textwidth]{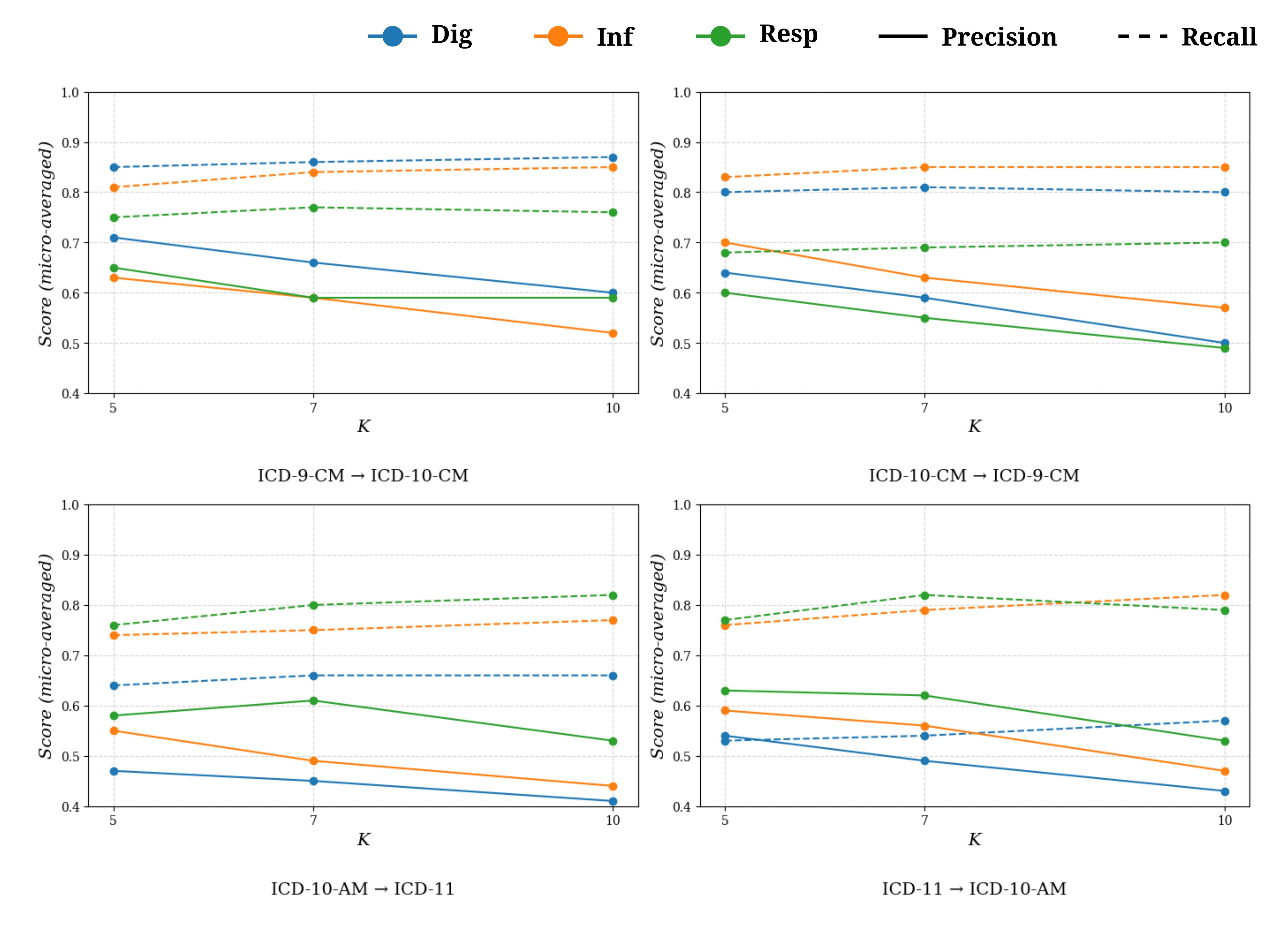}
        \caption*{(a) Micro-averaged}
    \end{subfigure}

     \begin{subfigure}{\textwidth}
        \centering
        \includegraphics[width=0.75\textwidth]{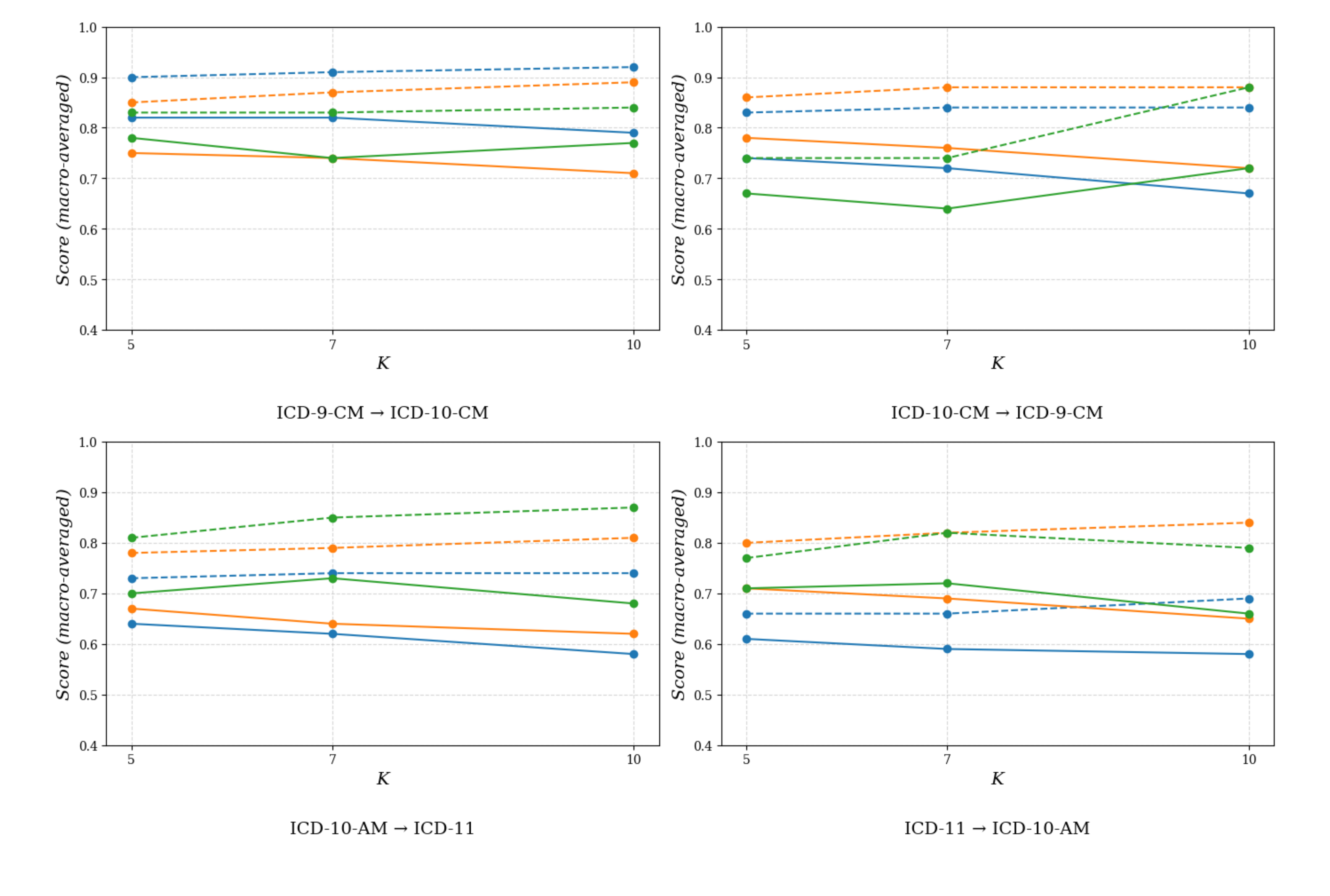}
        \caption*{(b) Macro-averaged}
    \end{subfigure}
    
    \caption{Effect of block size ($K$) on micro-averaged (top) and macro-averaged (bottom) precision (solid lines) and recall (dashed lines) for the blocking-and-matching pipeline across mapping pairs and chapters: Diseases of the Digestive System (\textbf{Dig}), Infectious and Parasitic Diseases (\textbf{Inf}), and Diseases of the Respiratory System (\textbf{Resp}) chapters. Increasing block size consistently reduces precision while maintaining or slightly improving recall.}
    \Description{Graphical comparison between the micro and macro averaged precision and recall values for different block size.}
    \label{fig:card_ablation}
\end{figure*}

Figure~\ref{fig:card_ablation} illustrates the effect of block size ($K$) on precision and recall obtained. Both micro-averaged and macro-averaged scores show similar patterns. As block size increases, precision often degrades while recall shows only minimal improvements. This highlights the critical importance of selecting an appropriate block size, as it affects not only mapping performance but also computational costs. Larger block sizes increase computation time (see Appendix~\ref{app:card_compute-time} for detailed comparisons) and, when using commercial LLMs such as OpenAI's ChatGPT\footnote{\url{https://chatgpt.com}}, incur higher costs since pricing is proportional to input token size. For large-scale mapping tasks, a block size of 10 would result in significantly higher costs compared to a block size of 5, with only marginal gains in recall and potential losses in precision.

\section{Prompt Example}
\label{app:prompt-eg}
Figure~\ref{fig:prompt-example} shows an example of our MCQ prompt, which uses the source code description as the anchor term and the candidate sets as the potential options and instructs the LLM to select all matching options (if any). It also includes the LLM response, including the final answer as well as the intermediate reasoning steps (as thinking content) used by the LLM.

\begin{figure*}[!ht]
    \begin{lstlisting}[
        basicstyle=\small\ttfamily,
    ]
    You are a medical coding expert tasked with identifying semantically equivalent codes.
    Source Code Description:
    late effects of viral encephalitis
    Select ALL options that represent the same or highly similar clinical condition as the source.
    B004 Herpesviral encephalitis
    A858 Other specified viral encephalitis
    B941 Sequelae of viral encephalitis
    Instructions:
    Return a **single Python-style list of strings**.
    - The list should contain only the matching `code` values.
    - If none of the options match, return `[]`.
    - **DO NOT include any explanation, additional text, or special characters outside the list.**
    
    <think>
    Okay, let's tackle this problem. The user wants to find semantically equivalent codes for "late effects of viral encephalitis" from the given options. Let me start by understanding each part.
    
    First, the source code description is "late effects of viral encephalitis." Late effects usually refer to complications or residual conditions that occur after the initial illness has resolved. In medical coding, these are often categorized under sequelae.
    
    Looking at the options:
    
    A.B004 Herpesviral encephalitis-This is an acute form of encephalitis caused by herpes viruses.The source is about late effects, not the acute phase, so this might not be a match.
    B. A858 Other specified viral encephalitis-This seems to be a general code for viral encephalitis that's not specified further.However, the source is about late effects, not the active disease. So maybe not directly related.
    C. B941 Sequelae of viral encephalitis-The term "sequelae" directly relates to late effects or complications after the initial illness.This seems like a strong match because it specifically refers to the aftermath of viral encephalitis, which aligns with the source description.
    
    So, the correct code should be B941. The other options are for the active disease (B004) or unspecified viral encephalitis (A858), which don't capture the late effects aspect. Therefore, the answer is [B941].
    </think>
    
    ["B941"]
    \end{lstlisting}
    \caption{Example of our MCQ prompt and the LLM's response, which include the reasoning steps (as the thinking content) and the final answer.}
    \label{fig:prompt-example}
    \Description{Prompt template used for LLM-based selector framework.}
\end{figure*}

\section{Precision-Recall-Coverage Trade-offs: Threshold-Based vs. Top-\textit{K} Selection Methods}
\label{app:precision-recall-tradeoff}

Threshold-based and top-$K$ selection methods present inherent trade-offs between precision, recall, and mapping coverage (the proportion of source codes that receive at least one target mapping). To illustrate these trade-offs, we evaluate both methods across varying parameter values.

\subsection{Threshold-Based Selection}

The threshold-based method prioritizes either precision or recall depending on the threshold value. Table~\ref{tab:card_th-tradeoffs} demonstrates this relationship across different threshold values. At $\lambda=0.95$, the method achieves high precision ($0.60-0.97$ in micro-averaged metrics) but suffers from low recall ($0.19-0.46$). Conversely, $\lambda=0.65$ maximizes recall ($0.85-0.97$) at the expense of precision ($0.04-0.20$), resulting in many low-confidence predictions. While macro-averaged results show more balanced performance than micro-averaged metrics, the overall trend remains consistent. This highlights the restrictive nature of high thresholds, which exclude cases where valid target codes score slightly below the cutoff. Conversely, lower thresholds improve both recall and coverage by accepting more candidate mappings, but significantly reduce precision due to the inclusion of false positives.The choice of threshold therefore depends on the application requirements, whether minimizing false positives (high threshold) or maximizing mapping completeness (low threshold) is more critical.

Furthermore, as shown in Table~\ref{tab:card_th-map-coverage}, the threshold value also affects the mapping coverage. For instance, at $\lambda=0.95$, coverage drops to $0.24-0.54$, meaning that over half of source codes produce no mappings above the threshold. While at $\lambda=0.65$, near-complete coverage ($0.93-1.00$) is achieved across all mapping directions and disease categories. These results demonstrate that threshold selection involves a three-way tradeoff among precision, recall, and coverage.

\begin{table*}[!ht]
\centering
\caption{Precision-Recall trade-offs for threshold-based selection methods using (a) micro-averaged and (b) macro-averaged methods. Results are shown for four threshold values ($\lambda \in \{0.95, 0.85, 0.75, 0.65\}$) across three disease categories: Diseases of the Digestive System (\textbf{Dig}), Infectious and Parasitic Diseases (\textbf{Inf}) and Diseases of the Respiratory System (\textbf{Resp}). Higher thresholds favour precision at the cost of recall, while lower thresholds maximize recall at the expense of precision.}
    \begin{subtable}[t]{0.48\linewidth}
        \resizebox{\linewidth}{!}{
            \begin{tabular}{llcccccc}
            \toprule
            & & \multicolumn{2}{c}{\textbf{Dig}} & \multicolumn{2}{c}{\textbf{Inf}} & \multicolumn{2}{c}{\textbf{Resp}} \\
            \cmidrule(lr){3-4} \cmidrule(lr){5-6} \cmidrule(lr){7-8}
            & & \textbf{Pr} & \textbf{Rc} & \textbf{Pr} & \textbf{Rc} & \textbf{Pr} & \textbf{Rc} \\
            \midrule
            \multirow{4.5}{*}{$\lambda=0.95$} & ICD-9-CM $\rightarrow$ ICD-10-CM & 0.66 & 0.46 & 0.89 & 0.25 & 0.77 & 0.40 \\
            & ICD-10-CM $\rightarrow$ ICD-9-CM & 0.60 & 0.39 & 0.97 & 0.22 & 0.80 & 0.31 \\
            & ICD-10-AM $\rightarrow$ ICD-11 & 0.81 & 0.28 & 0.82 & 0.29 & 0.82 & 0.46 \\
            & ICD-11 $\rightarrow$ ICD-10-AM & 0.84 & 0.19 & 0.86 & 0.31 & 0.92 & 0.45 \\
            \midrule
            \multirow{4.5}{*}{$\lambda=0.85$} & ICD-9-CM $\rightarrow$ ICD-10-CM & 0.19 & 0.82 & 0.41 & 0.63 & 0.39 & 0.66 \\
            & ICD-10-CM $\rightarrow$ ICD-9-CM & 0.16 & 0.71 & 0.52 & 0.60 & 0.39 & 0.56 \\
            & ICD-10-AM $\rightarrow$ ICD-11 & 0.42 & 0.58 & 0.40 & 0.67 & 0.39 & 0.75 \\
            & ICD-11 $\rightarrow$ ICD-10-AM & 0.37 & 0.49 & 0.43 & 0.69 & 0.47 & 0.69 \\
            \midrule
            \multirow{4.5}{*}{$\lambda=0.75$} & ICD-9-CM $\rightarrow$ ICD-10-CM & 0.09 & 0.91 & 0.18 & 0.86 & 0.15 & 0.82 \\
            & ICD-10-CM $\rightarrow$ ICD-9-CM & 0.08 & 0.84 & 0.24 & 0.85 & 0.15 & 0.76 \\
            & ICD-10-AM $\rightarrow$ ICD-11 & 0.17 & 0.77 & 0.20 & 0.86 & 0.16 & 0.87 \\
            & ICD-11 $\rightarrow$ ICD-10-AM & 0.18 & 0.72 & 0.21 & 0.85 & 0.17 & 0.77 \\
            \midrule
            \multirow{4.5}{*}{$\lambda=0.65$} & ICD-9-CM $\rightarrow$ ICD-10-CM & 0.04 & 0.97 & 0.07 & 0.95 & 0.06 & 0.95 \\
            & ICD-10-CM $\rightarrow$ ICD-9-CM & 0.04 & 0.94 & 0.09 & 0.94 & 0.06 & 0.93 \\
            & ICD-10-AM $\rightarrow$ ICD-11 & 0.06 & 0.88 & 0.08 & 0.94 & 0.05 & 0.97 \\
            & ICD-11 $\rightarrow$ ICD-10-AM & 0.09 & 0.85 & 0.09 & 0.95 & 0.06 & 0.86 \\
            \bottomrule
            \end{tabular}
        }
        \caption*{(a)}
    \end{subtable} \hfill
    \begin{subtable}[t]{0.48\linewidth}
    \resizebox{\linewidth}{!}{
        \begin{tabular}{llcccccc}
        \toprule
        & & \multicolumn{2}{c}{\textbf{Dig}} & \multicolumn{2}{c}{\textbf{Inf}} & \multicolumn{2}{c}{\textbf{Resp}} \\
        \cmidrule(lr){3-4} \cmidrule(lr){5-6} \cmidrule(lr){7-8}
        & & \textbf{Pr} & \textbf{Rc} & \textbf{Pr} & \textbf{Rc} & \textbf{Pr} & \textbf{Rc} \\
        \midrule
        \multirow{4.5}{*}{$\lambda=0.95$} & ICD-9-CM $\rightarrow$ ICD-10-CM & 0.47 & 0.51 & 0.27 & 0.27 & 0.43 & 0.46 \\
        & ICD-10-CM $\rightarrow$ ICD-9-CM & 0.35 & 0.42 & 0.24 & 0.24 & 0.33 & 0.35 \\
        & ICD-10-AM $\rightarrow$ ICD-11 & 0.34 & 0.34 & 0.31 & 0.32 & 0.47 & 0.50 \\
        & ICD-11 $\rightarrow$ ICD-10-AM & 0.24 & 0.24 & 0.33 & 0.33 & 0.44 & 0.45 \\
        \midrule
        \multirow{4.5}{*}{$\lambda=0.85$} & ICD-9-CM $\rightarrow$ ICD-10-CM & 0.48 & 0.86 & 0.45 & 0.66 & 0.48 & 0.73 \\
        & ICD-10-CM $\rightarrow$ ICD-9-CM & 0.43 & 0.73 & 0.46 & 0.62 & 0.41 & 0.60 \\
        & ICD-10-AM $\rightarrow$ ICD-11 & 0.50 & 0.67 & 46 & 0.71 & 0.50 & 0.79 \\
        & ICD-11 $\rightarrow$ ICD-10-AM & 0.37 & 0.51 & 0.47 & 0.70 & 0.47 & 0.69 \\
        \midrule
        \multirow{4.5}{*}{$\lambda=0.75$} & ICD-9-CM $\rightarrow$ ICD-10-CM & 0.30 & 0.88 & 0.31 & 0.88 & 0.31 & 0.88 \\
        & ICD-10-CM $\rightarrow$ ICD-9-CM & 0.30 & 0.85 & 0.39 & 0.85 & 0.25 & 0.78 \\
        & ICD-10-AM $\rightarrow$ ICD-11 & 0.35 & 0.83 & 0.34 & 0.88 & 0.33 & 0.90 \\
        & ICD-11 $\rightarrow$ ICD-10-AM & 0.34 & 0.71 & 0.34 & 0.85 & 0.29 & 0.77 \\
        \midrule
        \multirow{4.5}{*}{$\lambda=0.65$} & ICD-9-CM $\rightarrow$ ICD-10-CM & 0.12 & 0.98 & 0.14 & 0.96 & 0.13 & 0.97 \\
        & ICD-10-CM $\rightarrow$ ICD-9-CM & 0.12 & 0.94 & 0.20 & 0.95 & 0.14 & 0.94 \\
        & ICD-10-AM $\rightarrow$ ICD-11 & 0.14 & 0.92 & 0.17 & 0.95 & 0.14 & 0.98 \\
        & ICD-11 $\rightarrow$ ICD-10-AM & 0.19 & 0.86 & 0.20 & 0.95 & 0.13 & 0.86 \\
        \bottomrule
        \end{tabular}
    }
    \caption*{(b)}
    \end{subtable}
    \label{tab:card_th-tradeoffs}
\end{table*}

\begin{table}[ht]
\centering
\caption{Performance metrics for threshold-based selection across varying threshold values.}
\resizebox{0.75\linewidth}{!}{
\begin{tabular}{llccc}
\toprule
& & \textbf{Dig} & \textbf{Inf} & \textbf{Resp} \\
\midrule
\multirow{4.5}{*}{$\lambda=0.95$} & ICD-9-CM $\rightarrow$ ICD-10-CM & 0.54 & 0.29 & 0.50 \\
& ICD-10-CM $\rightarrow$ ICD-9-CM & 0.45 & 0.24 & 0.38 \\
& ICD-10-AM $\rightarrow$ ICD-11 & 0.41 & 0.37 & 0.53 \\
& ICD-11 $\rightarrow$ ICD-10-AM & 0.26 & 0.38 & 0.46 \\
\midrule
\multirow{4.5}{*}{$\lambda=0.85$} & ICD-9-CM $\rightarrow$ ICD-10-CM & 0.93 & 0.73 & 0.83 \\
& ICD-10-CM $\rightarrow$ ICD-9-CM & 0.78 & 0.68 & 0.71 \\
& ICD-10-AM $\rightarrow$ ICD-11 & 0.82 & 0.81 & 0.87 \\
& ICD-11 $\rightarrow$ ICD-10-AM & 0.57 & 0.77 & 0.72 \\
\midrule
\multirow{4.5}{*}{$\lambda=0.75$} & ICD-9-CM $\rightarrow$ ICD-10-CM & 0.98 & 0.94 & 0.94 \\
& ICD-10-CM $\rightarrow$ ICD-9-CM & 0.93 & 0.93 & 0.90 \\
& ICD-10-AM $\rightarrow$ ICD-11 & 0.95 & 0.97 & 0.97 \\
& ICD-11 $\rightarrow$ ICD-10-AM & 0.80 & 0.90 & 0.85 \\
\midrule
\multirow{4.5}{*}{$\lambda=0.65$} & ICD-9-CM $\rightarrow$ ICD-10-CM & 1.00 & 0.98 & 1.00 \\
& ICD-10-CM $\rightarrow$ ICD-9-CM & 0.99 & 0.99 & 0.99 \\
& ICD-10-AM $\rightarrow$ ICD-11 & 1.00 & 1.00 & 1.00 \\
& ICD-11 $\rightarrow$ ICD-10-AM & 0.96 & 0.98 & 0.93 \\
\bottomrule
\end{tabular}
}
\label{tab:card_th-map-coverage}
\end{table}

\subsection{Top-\texttt{K} Selection}

Unlike the threshold-based method, the top-$K$ selection method achieves a full mapping coverage; however, it prioritizes recall over precision (Table~\ref{tab:card_topk-tradeoffs}). At $K=3$, recall (micro-averaged) ranges from $0.64$ to $0.86$, substantially higher than the threshold-based method at comparable precision levels. As $K$ increases to 10, recall reaches $0.81$--$0.95$ while precision decreases to $0.09$--$0.11$. Similar trends hold for macro-averaged precision and recall values. This trade-off is inherent to the top-$K$ design: by guaranteeing exactly $K$ mappings per source code, the method maximizes coverage at the expense of prediction quality, making it suitable for applications that require comprehensive mapping completeness over precision.

\begin{table*}[!ht]
\centering
\caption{Precision-Recall trade-offs for top-$K$ selection methods using (a) micro-averaged and (b) macro-averaged methods. Results are shown for four different values for $K$ ($K \in \{3, 5, 7, 10\}$) across three disease categories: Diseases of the Digestive System (\textbf{Dig}), Infectious and Parasitic Diseases (\textbf{Inf}) and Diseases of the Respiratory System (\textbf{Resp}). Generally, the top-$K$ selection method prioritizes the recall over the precision.}
    \begin{subtable}[t]{0.48\linewidth}
        \resizebox{\linewidth}{!}{
            \begin{tabular}{llcccccc}
                \toprule
                & & \multicolumn{2}{c}{\textbf{Dig}} & \multicolumn{2}{c}{\textbf{Inf}} & \multicolumn{2}{c}{\textbf{Resp}} \\
                \cmidrule(lr){3-4} \cmidrule(lr){5-6} \cmidrule(lr){7-8}
                & & \textbf{Pr} & \textbf{Rc} & \textbf{Pr} & \textbf{Rc} & \textbf{Pr} & \textbf{Rc} \\
                \midrule
                \multirow{4.5}{*}{$K=3$} & ICD-9-CM $\rightarrow$ ICD-10-CM & 0.32 & 0.85 & 0.31 & 0.83 & 0.31 & 0.78 \\
                & ICD-10-CM $\rightarrow$ ICD-9-CM & 0.31 & 0.82 & 0.31 & 0.86 & 0.29 & 0.73 \\
                & ICD-10-AM $\rightarrow$ ICD-11 & 0.29 & 0.69 & 0.29 & 0.80 & 0.31 & 0.85 \\
                & ICD-11 $\rightarrow$ ICD-10-AM & 0.29 & 0.64 & 0.29 & 0.80 & 0.26 & 0.79 \\
                \midrule
                \multirow{4.5}{*}{$K=5$} & ICD-9-CM $\rightarrow$ ICD-10-CM & 0.21 & 0.90 & 0.20 & 0.88 & 0.20 & 0.83 \\
                & ICD-10-CM $\rightarrow$ ICD-9-CM & 0.20 & 0.87 & 0.20 & 0.91 & 0.19 & 0.80 \\
                & ICD-10-AM $\rightarrow$ ICD-11 & 0.19 & 0.74 & 0.19 & 0.87 & 0.19 & 0.89 \\
                & ICD-11 $\rightarrow$ ICD-10-AM & 0.19 & 0.71 & 0.19 & 0.88 & 0.17 & 0.84 \\
                \midrule
                \multirow{4.5}{*}{$K=7$} & ICD-9-CM $\rightarrow$ ICD-10-CM & 0.15 & 0.92 & 0.15 & 0.92 & 0.15 & 0.86 \\
                & ICD-10-CM $\rightarrow$ ICD-9-CM & 0.14 & 0.90 & 0.15 & 0.93 & 0.14 & 0.83 \\
                & ICD-10-AM $\rightarrow$ ICD-11 & 0.14 & 0.78 & 0.14 & 0.90 & 0.14 & 0.91 \\
                & ICD-11 $\rightarrow$ ICD-10-AM & 0.15 & 0.76 & 0.14 & 0.92 & 0.12 & 0.87 \\
                \midrule
                \multirow{4.5}{*}{$K=10$} & ICD-9-CM $\rightarrow$ ICD-10-CM & 0.11 & 0.94 & 0.11 & 0.94 & 0.11 & 0.90 \\
                & ICD-10-CM $\rightarrow$ ICD-9-CM & 0.10 & 0.91 & 0.10 & 0.95 & 0.10 & 0.87 \\
                & ICD-10-AM $\rightarrow$ ICD-11 & 0.10 & 0.81 & 0.10 & 0.93 & 0.10 & 0.93 \\
                & ICD-11 $\rightarrow$ ICD-10-AM & 0.11 & 0.83 & 0.10 & 0.95 & 0.09 & 0.90 \\
                \bottomrule
            \end{tabular}
        }
        \caption*{(a)}
    \end{subtable} \hfill
    \begin{subtable}[t]{0.48\linewidth}
    \resizebox{\linewidth}{!}{
        \begin{tabular}{llcccccc}
            \toprule
                & & \multicolumn{2}{c}{\textbf{Dig}} & \multicolumn{2}{c}{\textbf{Inf}} & \multicolumn{2}{c}{\textbf{Resp}} \\
                \cmidrule(lr){3-4} \cmidrule(lr){5-6} \cmidrule(lr){7-8}
                & & \textbf{Pr} & \textbf{Rc} & \textbf{Pr} & \textbf{Rc} & \textbf{Pr} & \textbf{Rc} \\
                \midrule
                \multirow{4.5}{*}{$K=3$} & ICD-9-CM $\rightarrow$ ICD-10-CM & 0.32 & 0.89 & 0.31 & 0.87 & 0.30 & 0.85 \\
                & ICD-10-CM $\rightarrow$ ICD-9-CM & 0.30 & 0.85 & 0.31 & 0.88 & 0.28 & 0.78 \\
                & ICD-10-AM $\rightarrow$ ICD-11 & 0.29 & 0.79 & 0.29 & 0.84 & 0.31 & 0.90 \\
                & ICD-11 $\rightarrow$ ICD-10-AM & 0.29 & 0.75 & 0.28 & 0.84 & 0.26 & 0.79 \\
                \midrule
                \multirow{4.5}{*}{$K=5$} & ICD-9-CM $\rightarrow$ ICD-10-CM & 0.21 & 0.94 & 0.20 & 0.91 & 0.20 & 0.90 \\
                & ICD-10-CM $\rightarrow$ ICD-9-CM & 0.20 & 0.90 & 0.10 & 0.93 & 0.19 & 0.83 \\
                & ICD-10-AM $\rightarrow$ ICD-11 & 0.19 & 0.83 & 0.19 & 0.91 & 0.19 & 0.94 \\
                & ICD-11 $\rightarrow$ ICD-10-AM & 0.19 & 0.81 & 0.19 & 0.91 & 0.17 & 0.84 \\
                \midrule
                \multirow{4.5}{*}{$K=7$} & ICD-9-CM $\rightarrow$ ICD-10-CM & 0.15 & 0.96 & 0.14 & 0.94 & 0.14 & 0.92 \\
                & ICD-10-CM $\rightarrow$ ICD-9-CM & 0.14 & 0.92 & 0.14 & 0.94 & 0.14 & 0.86 \\
                & ICD-10-AM $\rightarrow$ ICD-11 & 0.14 & 0.86 & 0.14 & 0.93 & 0.14 & 0.95 \\
                & ICD-11 $\rightarrow$ ICD-10-AM & 0.15 & 0.85 & 0.14 & 0.94 & 0.12 & 0.87 \\
                \midrule
                \multirow{4.5}{*}{$K=10$} & ICD-9-CM $\rightarrow$ ICD-10-CM & 0.11 & 0.97 & 0.11 & 0.96 & 0.11 & 0.95 \\
                & ICD-10-CM $\rightarrow$ ICD-9-CM & 0.10 & 0.94 & 0.10 & 0.96 & 0.10 & 89 \\
                & ICD-10-AM $\rightarrow$ ICD-11 & 0.10 & 0.89 & 0.10 & 0.95 & 0.10 & 0.97 \\
                & ICD-11 $\rightarrow$ ICD-10-AM & 0.11 & 0.88 & 0.10 & 0.95 & 0.09 & 0.90 \\
            \bottomrule
        \end{tabular}
    }
    \caption*{(b)}
    \end{subtable}
    \label{tab:card_topk-tradeoffs}
\end{table*}

\subsection{Comparative Analysis}

Comparing Tables~\ref{tab:card_th-tradeoffs} and~\ref{tab:card_topk-tradeoffs} reveals distinct trade-offs between precision, recall, and coverage. With the threshold-based method, the value of $\lambda$ directly influences precision, recall, and mapping coverage: higher thresholds favour precision but reduce coverage, while lower thresholds maximize coverage and recall at the expense of precision. The top-$K$ method, while guaranteeing complete mapping coverage, consistently prioritizes recall over precision, regardless of the $K$ value. This fundamental difference makes threshold-based selection suitable for high-confidence applications and top-$K$ selection appropriate for scenarios requiring comprehensive coverage.

As discussed in Section~\ref{sub-sec:block}, we use a combination of top-$K$ selection and \emph{bidirectional mapping expansion} in the blocking stage. This design choice is motivated by the observation that lower threshold values are required to attain full mapping coverage with the threshold-based method. However, low thresholds result in significantly larger block sizes, which reduces efficiency, a critical aspect of robust blocking methods. The top-$K$ approach addresses this limitation by ensuring complete coverage while maintaining controllable block sizes through the $K$ parameter, making it more suitable for the blocking stage where efficiency is crucial.

\section{Block Size Comparison: Threshold-Based vs. Top-\textit{K} Selection}
\label{app:card_block-size-comp}
As discussed in Section~\ref{sub-sec:block}, we selected top-$K$ over threshold-based selection for blocking. Table~\ref{tab:card_block-size-comp} demonstrates that top-$K$ ($K=5$) achieves comparable or superior recall (micro-averaged) to threshold-based selection ($\lambda=0.75$) while maintaining significantly smaller average block sizes across most mapping task. 

Top-$K$ ($K=5$) guarantees a fixed block size of 5.00 candidates per source code, while the threshold-based method produces variable block sizes ranging from 4.25 to 12.61 (average: 6.90). In 9 out of 12 mapping task, top-$K$ achieves equal or higher micro-averaged recall with smaller or equal block sizes. Therefore, we opted for top-$K$ in our hybrid blocking strategy.

\begin{table*}[!ht]
    \centering
    \caption{Micro-averaged recall (\textbf{Rc}) and average block size (\textbf{Avg. Size}) comparison between threshold-based ($\lambda=0.75$) and top-$K$ ($K=5$) selection methods. Bold values indicate better performance for each metric.}
    \resizebox{0.75\linewidth}{!}{
        \begin{tabular}{llcccccc}
            \toprule
            & & \multicolumn{2}{c}{\textbf{Dig}} & \multicolumn{2}{c}{\textbf{Inf}} & \multicolumn{2}{c}{\textbf{Resp}} \\
            \cmidrule(lr){3-4} \cmidrule(lr){5-6} \cmidrule(lr){7-8}
            & & \textbf{Rc}($\uparrow$) & \textbf{Avg. Size}($\downarrow$) & \textbf{Rc}($\uparrow$) & \textbf{Avg. Size}($\downarrow$) & \textbf{Rc}($\uparrow$) & \textbf{Avg. Size}($\downarrow$) \\
            \midrule
            \multirow{2.5}{*}{ICD-9-CM $\rightarrow$ ICD-10-CM} & Top-$K$($K=5$) & 0.90 & \textbf{5.00} & \textbf{0.88} & \textbf{5.00} & \textbf{0.83} & \textbf{5.00} \\
            & Threshold($\lambda=0.75$) & \textbf{0.91} & 12.11 & 0.86 & 5.59 & 0.82 & 6.95 \\
            \midrule
             \multirow{2.5}{*}{ICD-10-CM $\rightarrow$ ICD-9-CM} & Top-$K$($K=5$) & \textbf{0.87} & \textbf{5.00} & \textbf{0.91} & 5.00 & \textbf{0.80} & \textbf{5.00} \\
            & Threshold($\lambda=0.75$) & 0.84 & 12.61 & 0.85 & \textbf{4.25} & 0.76 & 6.43 \\
            \midrule
             \multirow{2.5}{*}{ICD-10-AM $\rightarrow$ ICD-11} & Top-$K$($K=5$) & 0.74 & \textbf{5.00} & \textbf{0.87} & 5.00 & \textbf{0.89} & \textbf{5.00} \\
            & Threshold($\lambda=0.75$) & \textbf{0.77} & 5.95 & 0.86 & \textbf{4.98} & 0.77 & 6.22 \\
            \midrule
            \multirow{2.5}{*}{ICD-11 $\rightarrow$ ICD-10-AM} & Top-$K$($K=5$) & 0.71 & \textbf{5.00} & \textbf{0.88} & 5.00 & \textbf{0.84} & \textbf{5.00} \\
            & Threshold($\lambda=0.75$) & \textbf{0.72} & 6.70 & 0.85 & \textbf{4.83} & 0.77 & 5.22 \\
            \bottomrule
        \end{tabular}
    }
    \label{tab:card_block-size-comp}
\end{table*}

\section{Effect of Top-\textit{K}+BiMaps Blocking Strategy on Block Size and Recall.}
\label{app:card_effect-hybrid-blocking-block-size}
In this section, we compare the simple top-$K$ selection approach and the hybrid blocking strategy (i.e., Top-$K$+BiMaps) in terms of the candidate set size and the overall recall.

Table~\ref{tab:card_effect-hybrid-block-size} shows the average block size for each source code across different mapping pairs and chapters. The block size remains very close to the corresponding $K$ values, suggesting that combining bidirectional mapping expansion and top-$K$ selection methods has minimal effect on the block size.

\begin{table*}[!htb]
	\centering
    \caption{Effect of the hybrid blocking strategy (Top-$K$+BiMaps)1 on candidate set size. Values represent the average number of candidate target codes per source code. The bidirectional mapping expansion has minimal impact, as average block sizes remain close to the corresponding $K$ values.}
    \resizebox{\linewidth}{!}{
	\begin{tabular}{lcccccccccccc}
		\toprule
	      \multirow{2.5}{*}{$K$}& \multicolumn{3}{c}{\textbf{ICD-9-CM $\rightarrow$ ICD-10-CM}} & \multicolumn{3}{c}{\textbf{ICD-10-CM $\rightarrow$ ICD-9-CM}} & \multicolumn{3}{c}{\textbf{ICD-10-AM $\rightarrow$ ICD-11}} & \multicolumn{3}{c}{\textbf{ICD-11 $\rightarrow$ ICD-10-AM}} \\
        
        \cmidrule(lr){2-4}\cmidrule(lr){5-7}\cmidrule(lr){8-10}\cmidrule(lr){11-13}
        
		& \textbf{Dig} & \textbf{Inf} & \textbf{Resp} & \textbf{Dig} & \textbf{Inf} & \textbf{Resp} & \textbf{Dig} & \textbf{Inf} & \textbf{Resp} & \textbf{Dig} & \textbf{Inf} & \textbf{Resp} \\
        
		\midrule
        $K=3$ & 3.37 & 3.31 & 3.34 & 3.24 & 3.11 & 3.14 & 3.85 & 3.32 & 3.29 & 3.11 & 3.14 & 3.09 \\
        $K=5$ & 5.25 & 5.19 & 5.26 & 5.17 & 5.06 & 5.10 & 5.61 & 5.23 & 5.22 & 5.08 & 5.08 & 5.06 \\
        $K=7$ & 7.19 & 7.13 & 7.20 & 7.14 & 7.04 & 7.07 & 7.49 & 7.18 & 7.20 & 7.05 & 7.04 & 7.05 \\ 
        \bottomrule
    \end{tabular}
    }
    \label{tab:card_effect-hybrid-block-size}
\end{table*}

Similarly, the hybrid blocking strategy consistently outperforms the top-$K$ baseline ($K=5$) across all mapping scenarios (Tables~\ref{tab:card_effect-hybrid-recall-micro} and~\ref{tab:card_effect-hybrid-recall-macro}). For micro-averaged recall, improvements range from 0.01 to 0.04, with the most substantial gains in ICD-10-CM $\rightarrow$ ICD-9-CM \textbf{Inf} (+0.03) and ICD-9-CM $\rightarrow$ ICD-10-CM \textbf{Resp} (+0.03) mappings. The macro-averaged results show similar patterns, with the hybrid method achieving equal or higher recall in all twelve scenarios, matching baseline performance in only two cases. Crucially, these recall gains are achieved with minimal efficiency cost (Table~\ref{tab:card_effect-hybrid-block-size}), demonstrating that bidirectional expansion captures additional matches without significantly increasing the context burden on the subsequent LLM-based matching stage.

\begin{table*}[!htb]
	\centering
    \caption{Micro-averaged recall comparison between top-$K$ baseline and Top-$K$+BiMaps blocking strategy ($K=5$) across different ICD mapping tasks and chapters. The hybrid approach consistently achieves higher recall across all twelve mapping scenarios with minimal increase in candidate set size.}
    \resizebox{0.80\linewidth}{!}{
	\begin{tabular}{lcccccccccccc}
		\toprule
	      \multirow{2}{*}{}& \multicolumn{3}{c}{\textbf{ICD-9-CM $\rightarrow$ ICD-10-CM}} & \multicolumn{3}{c}{\textbf{ICD-10-CM $\rightarrow$ ICD-9-CM}} & \multicolumn{3}{c}{\textbf{ICD-10-AM $\rightarrow$ ICD-11}} & \multicolumn{3}{c}{\textbf{ICD-11 $\rightarrow$ ICD-10-AM}} \\
        
        \cmidrule(lr){2-4}\cmidrule(lr){5-7}\cmidrule(lr){8-10}\cmidrule(lr){11-13}
        
		& \textbf{Dig} & \textbf{Inf} & \textbf{Resp} & \textbf{Dig} & \textbf{Inf} & \textbf{Resp} & \textbf{Dig} & \textbf{Inf} & \textbf{Resp} & \textbf{Dig} & \textbf{Inf} & \textbf{Resp} \\
        
		\midrule
        Top-$K$ & 0.90 & 0.88 & 0.83 & 0.87 & 0.91 & 0.80 & 0.74 & 0.87 & 0.89 & 0.71 & 0.88 & 0.84 \\ 
        Top-$K$+BiMaps & \textbf{0.92} & \textbf{0.90} & \textbf{0.86} & \textbf{0.88} & \textbf{0.92} & \textbf{0.82} & \textbf{0.74} & \textbf{0.89} & \textbf{0.90} & \textbf{0.72} & \textbf{0.89} & \textbf{0.85} \\
        \bottomrule
    \end{tabular}
    }
    \label{tab:card_effect-hybrid-recall-micro}
\end{table*}

\begin{table*}[!ht]
	\centering
    \caption{Macro-averaged recall comparison between top-$K$ baseline and Top-$K$+BiMaps blocking strategy ($K=5$) across different ICD mapping tasks and chapters. The Top-$K$+BiMaps approach achieves higher or equal recall in all but two scenarios (underlined), where both methods yield identical results.}
    \resizebox{0.80\linewidth}{!}{
	\begin{tabular}{lcccccccccccc}
		\toprule
	      \multirow{2}{*}{}& \multicolumn{3}{c}{\textbf{ICD-9-CM $\rightarrow$ ICD-10-CM}} & \multicolumn{3}{c}{\textbf{ICD-10-CM $\rightarrow$ ICD-9-CM}} & \multicolumn{3}{c}{\textbf{ICD-10-AM $\rightarrow$ ICD-11}} & \multicolumn{3}{c}{\textbf{ICD-11 $\rightarrow$ ICD-10-AM}} \\
        
        \cmidrule(lr){2-4}\cmidrule(lr){5-7}\cmidrule(lr){8-10}\cmidrule(lr){11-13}
        
		& \textbf{Dig} & \textbf{Inf} & \textbf{Resp} & \textbf{Dig} & \textbf{Inf} & \textbf{Resp} & \textbf{Dig} & \textbf{Inf} & \textbf{Resp} & \textbf{Dig} & \textbf{Inf} & \textbf{Resp} \\
        
		\midrule
        Top-$K$ & 0.94 & 0.91 & 0.90 & \underline{0.90} & 0.93 & 0.83 & 0.83 & 0.91 & \underline{0.94} & 0.81 & 0.91 & 0.84 \\
        Top-$K$+BiMaps & \textbf{0.95} & \textbf{0.93} & \textbf{0.91} & \underline{0.90} & \textbf{0.94} & \textbf{0.85} & \textbf{0.84} & \textbf{0.92} & \underline{0.94} & \textbf{0.82} & \textbf{0.92} & \textbf{0.85} \\
        \bottomrule
    \end{tabular}
    }
    \label{tab:card_effect-hybrid-recall-macro}
\end{table*}

\section{Computational Time for Blocking-and-Matching Pipeline}
\label{app:card_compute-time}
We ran all our experiments on a single NVIDIA A30 GPU node with 24GB of video memory. We set \verb|batch_size=8|. Table~\ref{tab:card_compute-time} shows the chapter-wise execution time (in hh:mm:ss format) required to complete each mapping task for different values of $K \in \{5, 7, 10\}$. As expected, larger block sizes generally result in longer computational times due to the increased number of candidate pairs requiring LLM-based evaluation. Compared to $K=5$ (total time: 43:38:52), $K=7$ required an additional 6:12:34, while $K=10$ required an additional 9:27:49 to complete all mappings.

\begin{table*}[!ht]
    \centering
    \caption{Execution time for each ICD mapping task across different mapping directions and values of $K$, reported in \textit{hh:mm:ss} format.}
    \resizebox{0.5\linewidth}{!}{
    \begin{tabular}{llccc}
        \toprule
         & & \textbf{$K=5$} & \textbf{$K=7$} & \textbf{$K=10$} \\
        \midrule
        \multirow{3}{*}{\textbf{ICD-9-CM $\rightarrow$ ICD-10-CM}} & Dig  & 3:45:17 & 3:48:30 & 4:00:14 \\
        & Inf  & 6:27:40 & 7:24:03 & 7:50:25 \\
        & Resp & 1:29:19 & 2:14:33 & 1:51:26 \\
        \midrule
        \multirow{3}{*}{\textbf{ICD-10-CM $\rightarrow$ ICD-9-CM}} & Dig  & 4:08:33 & 4:51:14 & 5:17:22 \\
        & Inf  & 6:53:30 & 7:20:39 & 8:11:22 \\
        & Resp & 2:23:51 & 2:44:41 & 2:45:13 \\
        \midrule
        \multirow{3}{*}{\textbf{ICD-10-AM $\rightarrow$ ICD-11}} & Dig  & 2:09:33 & 2:31:35 & 2:28:27 \\
        & Inf  & 5:50:58 & 6:40:34 & 6:59:50 \\
        & Resp & 1:20:02 & 1:34:04 & 1:48:24 \\
        \midrule
        \multirow{3}{*}{\textbf{ICD-11 $\rightarrow$ ICD-10-AM}} & Dig  & 3:47:27 & 4:15:15 & 4:16:00 \\
        & Inf  & 4:02:49 & 4:56:17 & 5:41:03 \\
        & Resp & 1:19:53 & 1:27:01 & 1:56:55 \\
        \midrule
        \multicolumn{2}{l}{\textbf{Total}} & 43:38:52 & 49:51:26 & 53:06:41 \\
        \bottomrule
    \end{tabular}
    }
    \label{tab:card_compute-time}
\end{table*}

\section{Granularity Mismatch Analysis}
\label{app:card_granularity_mismatch}

Table~\ref{tab:card_granularity_mismatch} presents a detailed breakdown of false positives that correspond to hierarchically valid parent codes of the ground-truth mappings, across all mapping tasks and disease chapters. 

The proportion of hierarchically valid parent codes varied across mapping tasks, with ICD-9-CM $\rightarrow$ ICD-10-CM showing the highest rates (24.66\%--36.57\%) and ICD-11 $\rightarrow$ ICD-10-AM showing the lowest (11.30\%--17.70\%). While such parent-level mappings are semantically valid and clinically meaningful, capturing the broader diagnostic category, they were scored as false positives under our evaluation criteria.

\begin{table*}[!htbp]
    \centering
    \caption{Proportion of cases where at least one predicted target code is a hierarchically valid parent target code, by mapping task and chapter. Numbers in parentheses indicate the total number of source codes for which the blocking-and-matching framework predicted at least one parent target code.}
    \resizebox{0.5\linewidth}{!}{
    \begin{tabular}{llc}
         \toprule
         \textbf{SRC $\rightarrow$ TGT} & \textbf{Chapter} & \textbf{Proportion (\%)} \\
         \midrule
         \multirow{3}{*}{ICD-9-CM $\rightarrow$ ICD-10-CM} & Dig (184) & 32.07 \\
         & Inf (309) & 36.57 \\
         & Resp (73) & 24.66 \\
         \midrule
         \multirow{3}{*}{ICD-10-CM $\rightarrow$ ICD-9-CM} & Dig (567) & 18.69 \\
         & Inf (--) & -- \\
         & Resp (256) & 20.31 \\
         \midrule
         \multirow{3}{*}{ICD-10-AM $\rightarrow$ ICD-11} & Dig (273) & 16.85 \\
         & Inf (428) & 30.14 \\
         & Resp (169) & 25.44 \\
         \midrule
         \multirow{3}{*}{ICD-11 $\rightarrow$ ICD-10-AM} & Dig (508) & 12.40 \\
         & Inf (548) & 17.70 \\
         & Resp (177) & 11.30 \\
         \bottomrule
    \end{tabular}
    }
    \label{tab:card_granularity_mismatch}
\end{table*}

\section{Performance Comparison of Baseline Methods, Bidirectional Mapping Expansion and the Blocking and Matching Framework Using Macro-Averaged Precision, Recall and F1 Scores.}
\label{app:card_macro-avg-results}

Table~\ref{tab:card_macro_results} shows the performance comparison between the chosen baselines and our proposed methods (BiMaps and Blocking-and-Matching framework) using macro-averaged precision, recall and F1-scores. This scores show similar patterns with the micro-averaged scores presented in Table~\ref{tab:card_main_results}.

\begin{table*}[!htbp]
    \centering
    \caption{Performance comparison of baseline selection methods, bidirectional mapping expansion, and the blocking-and-matching pipeline across four mapping tasks and three disease chapters: Diseases of the Digestive System (\textbf{Dig}), Infectious and Parasitic Diseases (\textbf{Inf}), and Diseases of the Respiratory System (\textbf{Resp}). The reported numbers are macro-averaged precision (\textbf{Pr}), recall (\textbf{Rc}), and F1-score (\textbf{F1}).}
    
    \begin{subtable}[t]{0.48\linewidth}
        \centering
        \resizebox{\linewidth}{!}{
            \begin{tabular}{lcccccccccccc}
                \toprule
                & \multicolumn{3}{c}{\textbf{Dig}} & \multicolumn{3}{c}{\textbf{Inf}} & \multicolumn{3}{c}{\textbf{Resp}} \\
                \cmidrule(lr){2-4} \cmidrule(lr){5-7} \cmidrule(lr){8-10}
                & \textbf{Pr} & \textbf{Rc} & \textbf{F1} & \textbf{Pr} & \textbf{Rc} & \textbf{F1} & \textbf{Pr} & \textbf{Rc} & \textbf{F1} \\
                \midrule
                Threshold ($\lambda=0.85$) & 0.48 & 0.86 & 0.62 & 0.45 & 0.66 & 0.54 & 0.48 & 0.73 & 0.58 \\
                Top-$K$ ($K=5$) & 0.21 & \textbf{0.94} & 0.34 & 0.20 & \textbf{0.91} & 0.33 & 0.20 & \textbf{0.90} & 0.33 \\
                \midrule
                BiMaps & 0.75 & 0.88 & 0.81 & 0.67 & 0.81 & 0.73 & \textbf{0.71} & 0.79 & \textbf{0.75}\\
                \midrule
                Blocking-and-Matching & \textbf{0.82} & 0.90 & \textbf{0.86} & \textbf{0.75} & 0.85 & \textbf{0.80} & 0.65 & 0.75 & 0.70 \\
                \bottomrule
            \end{tabular}
        }
        \caption{ICD-9-CM $\rightarrow$ ICD-10-CM}
    \end{subtable}\hfill
    \begin{subtable}[t]{0.48\linewidth}
        \centering
        \resizebox{\linewidth}{!}{
            \begin{tabular}{lcccccccccccc}
                \toprule
                & \multicolumn{3}{c}{\textbf{Dig}} & \multicolumn{3}{c}{\textbf{Inf}} & \multicolumn{3}{c}{\textbf{Resp}} \\
                \cmidrule(lr){2-4} \cmidrule(lr){5-7} \cmidrule(lr){8-10}
                & \textbf{Pr} & \textbf{Rc} & \textbf{F1} & \textbf{Pr} & \textbf{Rc} & \textbf{F1} & \textbf{Pr} & \textbf{Rc} & \textbf{F1} \\
                \midrule
                Threshold ($\lambda=0.85$) & 0.43 & 0.73 & 0.54 & 0.46 & 0.62 & 0.53 & 0.41 & 0.6 & 0.49 \\
                Top-$K$ ($K=5$) & 0.20 & \textbf{0.90} & 0.33 & 0.20 & \textbf{0.93} & 0.33 & 0.19 & \textbf{0.83} & 0.31 \\
                \midrule
                BiMaps & 0.67 & 0.76 & 0.71 & 0.75 & 0.80 & 0.77 & 0.62 & 0.68 & 0.65 \\
                \midrule
                Blocking-and-Matching & \textbf{0.74} & {0.83} & \textbf{0.78} & \textbf{0.78} & 0.86 & \textbf{0.82} & \textbf{0.67} & 0.74 & \textbf{0.70} \\
                \bottomrule
            \end{tabular}
        }
        \caption{ICD-10-CM $\rightarrow$ ICD-9-CM}
    \end{subtable}
    
    \vspace{0.01\linewidth}
    
    \begin{subtable}[t]{0.48\linewidth}
        \centering
        \resizebox{\linewidth}{!}{
            \begin{tabular}{lcccccccccccc}
                \toprule
                & \multicolumn{3}{c}{\textbf{Dig}} & \multicolumn{3}{c}{\textbf{Inf}} & \multicolumn{3}{c}{\textbf{Resp}} \\
                \cmidrule(lr){2-4} \cmidrule(lr){5-7} \cmidrule(lr){8-10}
                & \textbf{Pr} & \textbf{Rc} & \textbf{F1} & \textbf{Pr} & \textbf{Rc} & \textbf{F1} & \textbf{Pr} & \textbf{Rc} & \textbf{F1} \\
                \midrule
                Threshold ($\lambda=0.85$) & 0.5 & 0.67 & 0.57 & 0.46 & 0.71 & 0.56 & 0.5 & 0.79 & 0.61 \\
                Top-$K$ ($K=5$) & 0.19 & \textbf{0.83} & 0.31 & 0.19 & \textbf{0.91} & 0.31 & 0.19 & \textbf{0.94} & 0.32\\
                \midrule
                BiMaps & 0.58 & 0.74 & 0.65 & 0.64 & 0.72 & 0.68 & 0.69 & 0.79 & 0.74 \\
                \midrule
                Blocking-and-Matching & \textbf{0.64} & 0.73 & \textbf{0.68} & \textbf{0.67} & 0.78 & \textbf{0.72} & \textbf{0.70} & 0.81 & \textbf{0.75} \\
                \bottomrule
            \end{tabular}
        }
        \caption{ICD-10-AM $\rightarrow$ ICD-11}
    \end{subtable}\hfill
    \begin{subtable}[t]{0.48\linewidth}
        \centering
        \resizebox{\linewidth}{!}{
            \begin{tabular}{lcccccccccccc}
                \toprule
                & \multicolumn{3}{c}{\textbf{Dig}} & \multicolumn{3}{c}{\textbf{Inf}} & \multicolumn{3}{c}{\textbf{Resp}} \\
                \cmidrule(lr){2-4} \cmidrule(lr){5-7} \cmidrule(lr){8-10}
                & \textbf{Pr} & \textbf{Rc} & \textbf{F1} & \textbf{Pr} & \textbf{Rc} & \textbf{F1} & \textbf{Pr} & \textbf{Rc} & \textbf{F1} \\
                \midrule
                Threshold ($\lambda=0.85$) & 0.37 & 0.51 & 0.43 & 0.47 & 0.7 & 0.56 & 0.47 & 0.69 & 0.56\\
                Top-$K$ ($K=5$) & 0.19 & \textbf{0.81} & 0.31 & 0.19 & \textbf{0.91} & 0.31 & 0.17 & \textbf{0.84} & 0.28 \\
                \midrule
                BiMaps & \textbf{0.64} & 0.65 & \textbf{0.64} & 0.69 & 0.75 & 0.72 & 0.68 & 0.74 & 0.71 \\
                \midrule
                Blocking-and-Matching & 0.61 & 0.66 & 0.63 & \textbf{0.71} & 0.8 & \textbf{0.75} & \textbf{0.71} & 0.77 & \textbf{0.74} \\
                \bottomrule
            \end{tabular}
        }
        \caption{ICD-11 $\rightarrow$ ICD-10-AM}
    \end{subtable}
    \label{tab:card_macro_results}
\end{table*}

\end{document}